\documentclass{article}

\PassOptionsToPackage{numbers, sort&compress}{natbib}

\usepackage[final]{neurips_2024}

\usepackage[utf8]{inputenc} %
\usepackage[T1]{fontenc}    %
\usepackage{hyperref}       %
\usepackage{url}            %
\usepackage{booktabs}       %
\usepackage{amsfonts}       %
\usepackage{nicefrac}       %
\usepackage{microtype}      %
\usepackage{xcolor}         %
\usepackage{graphicx}
\usepackage{subcaption}
\usepackage{amsmath}
\usepackage{cleveref}
\usepackage{enumitem}
\usepackage{float}
\setlist[enumerate]{
  topsep=2pt,        %
  partopsep=0pt,     %
  itemsep=1pt,       %
  parsep=1pt,        %
  leftmargin=1.5em,  %
  rightmargin=0pt,   %
  listparindent=0pt, %
  itemindent=0pt,    %
}

\title{Characterizing stable regions\\in the residual stream of LLMs}

\author{
  Jett Janiak \\
  LASR Labs \\
  \texttt{jettjaniak@} \\
  \texttt{gmail.com} \\
  \And
  Jacek Karwowski$^*$ \\
  University of Oxford \\
  \texttt{jacek.karwowski@} \\
  \texttt{cs.ox.ac.uk}
  \And
  Chatrik Singh Mangat$^*$ \\
  LASR Labs\\
  \texttt{chatrikmangat@}\\
  \texttt{outlook.com}
  \AND
  Giorgi Giglemiani \\
  LASR Labs\\
  \texttt{giglemiani@} \\
  \texttt{gmail.com} \\
  \And
  Nora Petrova \\
  LASR Labs\\
  \texttt{nora.axion@} \\
  \texttt{gmail.com} \\
  \And
  Stefan Heimersheim \\
  Apollo Research\\
  \texttt{stefan@} \\
  \texttt{apolloresearch.ai}
}

\begin{document}

\maketitle
\def\thefootnote{*}\footnotetext{equal contribution}
\def\thefootnote{\arabic{footnote}}
\begin{abstract}
We identify \emph{stable regions} in the residual stream of Transformers, where the model's output remains insensitive to small activation changes, but exhibits high sensitivity at region boundaries. These regions emerge during training and become more defined as training progresses or model size increases. The regions appear to be much larger than previously studied polytopes. Our analysis suggests that these stable regions align with semantic distinctions, where similar prompts cluster within regions, and activations from the same region lead to similar next token predictions. This work provides a promising research direction for understanding the complexity of neural networks, shedding light on training dynamics, and advancing interpretability. 
\end{abstract}

\section{Introduction}
We study the effects of perturbing Transformer activations, building upon previous work \citep{gurnee_sae_2024, lindsey_how_2024, heimersheim_activation_2024}. Specifically, we interpolate between different residual stream activations after the first layer, and measure the change in the model output. Our initial results suggest that:
\begin{enumerate}
    \item The residual stream of a trained Transformer can be divided into \emph{stable regions}. Within these regions, small changes in model activations lead to minimal changes in output. However, at region boundaries, small changes can lead to significant output differences.
    
    \item These regions emerge during training and evolve with model scale. Randomly initialized models do not exhibit these stable regions, but as training progresses or model size increases, the boundaries between regions become sharper.
    
    \item These stable regions appear to correspond to \emph{semantic distinctions}. Dissimilar prompts occupy different regions, and activations from different regions produce different next token predictions.
    
    \item These stable regions appear to be much larger than previously studied polytopes \citep{montufar_number_2014, hanin_complexity_2019,black_interpreting_2022}. Our analysis of gate activations shows that while interpolating between two prompts typically crosses only one stable region boundary, hundreds or thousands of gates change their activation sign (\Cref{appendixF}).
\end{enumerate}

\begin{figure}
    \centering
    \includegraphics[width=\textwidth]{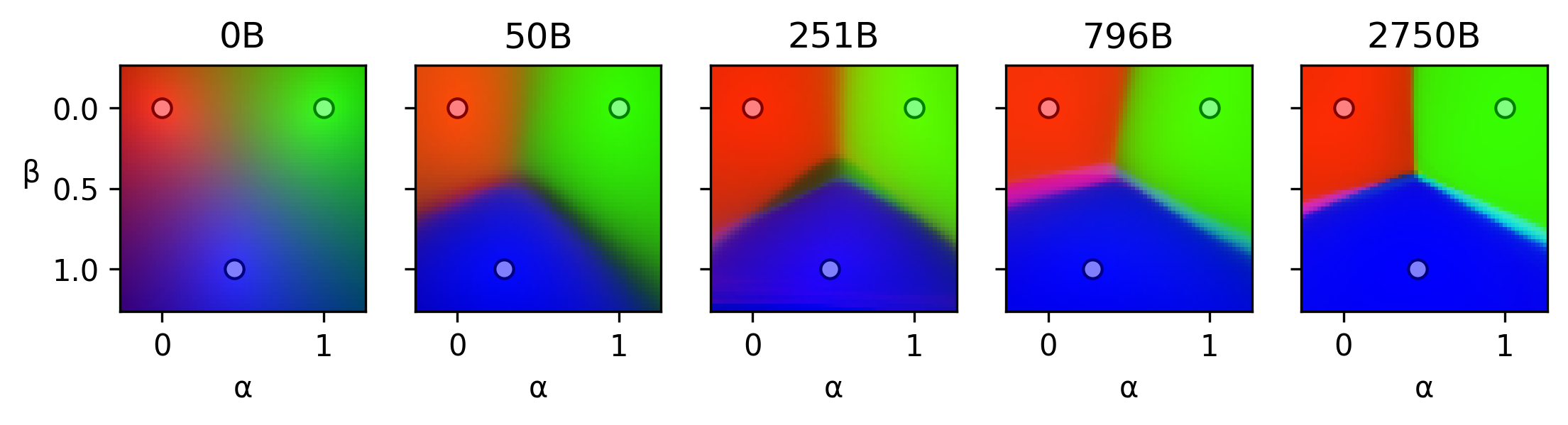}
    \caption{Visualization of stable regions in \texttt{OLMo-7B} during training. Colors represent the similarity of model outputs to those produced by three model-generated activations (red, green, blue circles). Each subplot shows a 2D slice of the residual stream after the first layer at different stages of training, with the number of processed tokens indicated in the titles. As training progresses from left to right, distinct regions of solid color emerge and the boundaries between them sharpen. Refer to the end of \Cref{sec:experiments} for details.}
    \label{fig:newone}
\end{figure}

\section{Related work}

Previous work on characterizing the complexity of neural networks includes studying linear regions (aka polytopes) in piecewise linear networks \citep{montufar_number_2014, hanin_complexity_2019, black_interpreting_2022} and bounding the VC dimension of network architectures \citep{bartlett_vapnik-chervonenkis_2003,harvey_nearly-tight_2017}. While these approaches provide theoretical insights, the extent to which they describe the practical expressivity of networks remains unclear. In this work, we study stable regions in the residual stream of Transformers that can be mapped to classes of very similar outputs. Our initial results suggest that stable regions are much larger than previously studies polytopes.

The linear representation hypothesis \citep{elhage_toy_2022, olah_what_2024, smith_strong_2024} suggests that features in neural networks correspond to directions in activation space. Since residual stream activations can be though of as sums of features, we can similarly think about model-generated activations as vectors. Due to normalization layers in Transformers, these vectors lie closely to a hypersphere in the residual stream. Because of high dimensionality of this space, most linear interpolations between two activations will stay close to the hypersphere.

\citet{heimersheim_activation_2024}, motivated by theoretical results by \citet{hanni_mathematical_2024}, studied robustness to noise in GPT-2 by interpolating between two residual stream activations and they observed nonlinear changes in the model output. Follow up studies by \citet{lee_investigating_2024, giglemiani_evaluating_2024} connect this to SAEs work by \citet{lindsey_how_2024,gurnee_sae_2024}.

\section{Methods}\label{methods}

For a given prompt $p$, we define a Transformer forward pass with activation patching as a function $F_p: \mathbb{R}^D \rightarrow \mathbb{R}^D$ that returns residual stream activations after last layer\footnote{Before the unembedding and the softmax.}, at last sequence position. Specifically, $F_p(X)$ replaces the residual stream activations after the first\footnote{Preliminary results in \Cref{appendixE} suggest that some of the reported effects hold for later layers.} layer, at the last sequence position, with $X$, before applying the rest of the model layers.

If small changes in model activations lead to minimal changes in output, then the residual stream after the first layer can be divided into stable regions. In this case, $F_p$'s rate of change should be small away from region boundaries and large when approaching and crossing region boundaries.

To study this efficiently, we interpolate between prompts $p_A$ and $p_B$, which may or may not belong to the same region. We introduce an interpolation coefficient $\alpha \in [0,1]$, where $\alpha = 0$ corresponds to prompt $p_A$ and $\alpha = 1$ corresponds to prompt $p_B$. 

We measure the $L_2$ distance $d(\alpha): [0,1] \rightarrow \mathbb{R}$ between two outputs: a clean run on prompt $p_A$ and a patched run where activations are modified depending on $\alpha$. In the patched run, we replace the activations with a linear interpolation $A + \alpha(B - A)$, where $A$ and $B$ represent the model-generated activations after the first layer, at the last sequence position, corresponding to prompts $p_A$ and $p_B$ respectively. Formally, this distance is expressed as:
\[
 d(\alpha) = \| F_{p_A}(A) - F_{p_A}(A + \alpha(B - A)) \|_2.
\]

In our experiments, we use models from $\texttt{Qwen2}$~\citep{yang_qwen2_2024} and $\texttt{OLMo}$~\citep{groeneveld_olmo_2024} families with a number of parameters ranging from $0.5$B to $7$B, see~\Cref{appendixA} for details. We compute the distance $d$ for 50 uniformly spaced values of $\alpha$ between 0 and 1.

\section{Experiments}\label{sec:experiments}

\paragraph{Illustrative examples}\label{ex_illustrative}
We suspect that semantically similar prompts are more likely to belong to the same stable region than semantically different prompts. To test this, we construct pairs of prompts dissimilar to each other numbered D1-D3, and pairs of similar prompts numbered S1-S3. We present prompts D1 and S1 below, and the remaining prompts in~\Cref{appendixB}. For simplicity, the prompts differ only in the last token.\\
Dissimilar prompts D1:\\
$p_A$ = `` The house at the end of the street was very'', top prediction = `` quiet''\\
$p_B$ = `` The house at the end of the street was in'', top prediction = `` a''\\
Similar prompts S1:\\
$p_A$ = `` She opened the dusty book and a cloud of mist''\\
$p_B$ = `` She opened the dusty book and a cloud of dust''

\begin{figure}%
    \centering
    \begin{subfigure}{0.3307\textwidth}
    \includegraphics[width=\textwidth]{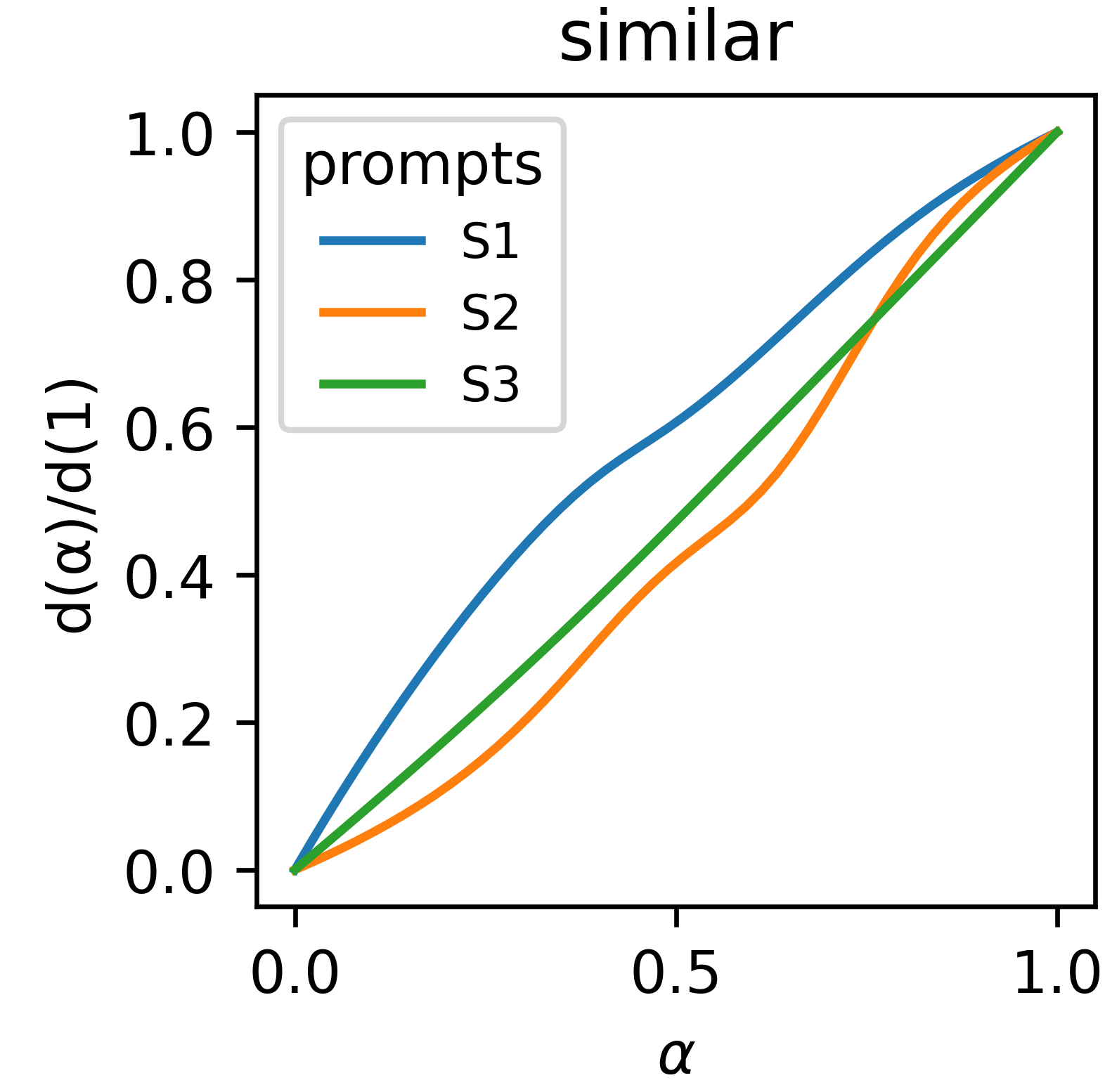}
    \caption{}\label{fig:onea}
    \end{subfigure}
    \begin{subfigure}{0.2692\textwidth}
    \includegraphics[width=\textwidth]{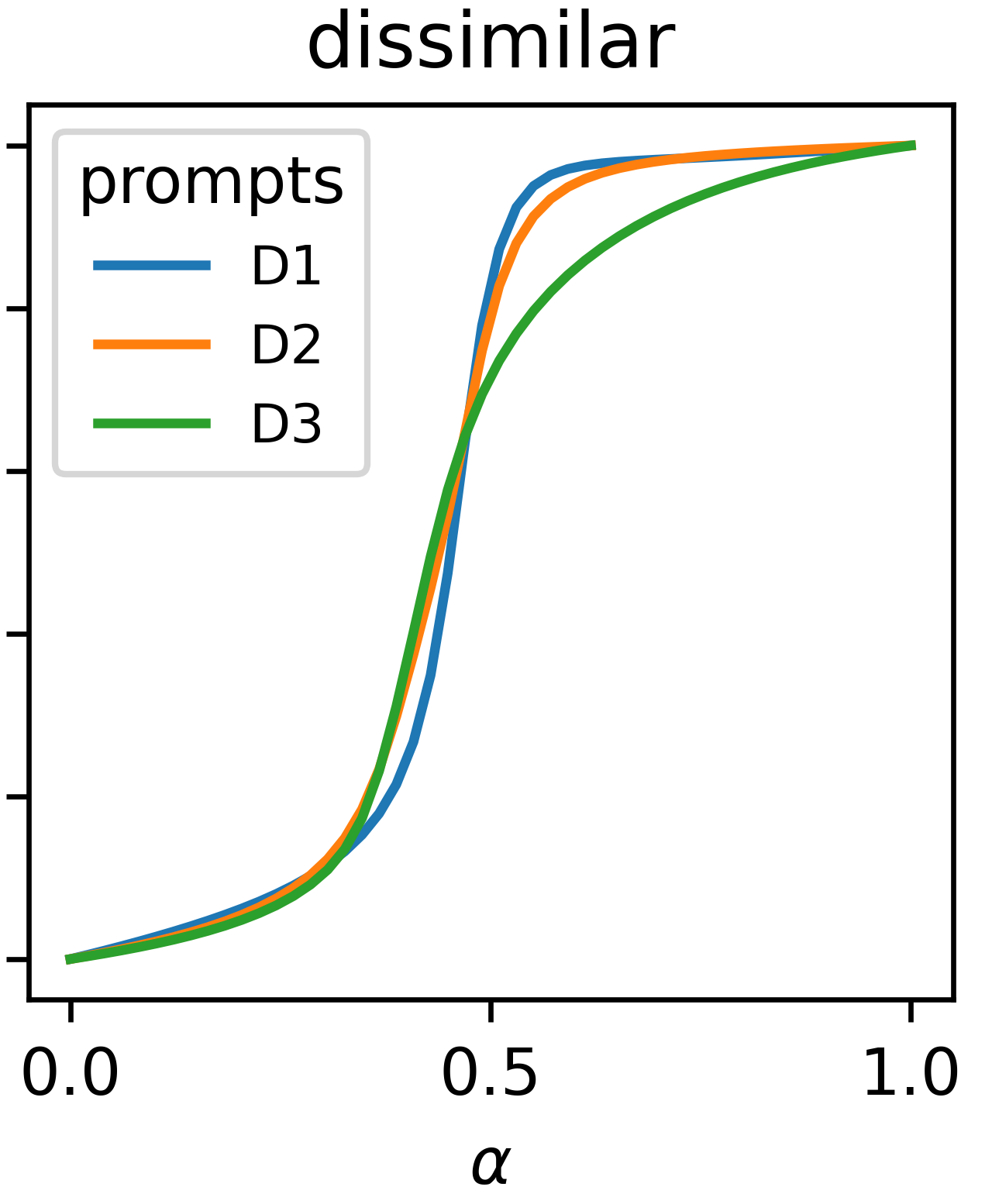}
    \caption{}\label{fig:oneb}
    \end{subfigure}
    \begin{subfigure}{0.3401\textwidth}
    \includegraphics[width=\textwidth]{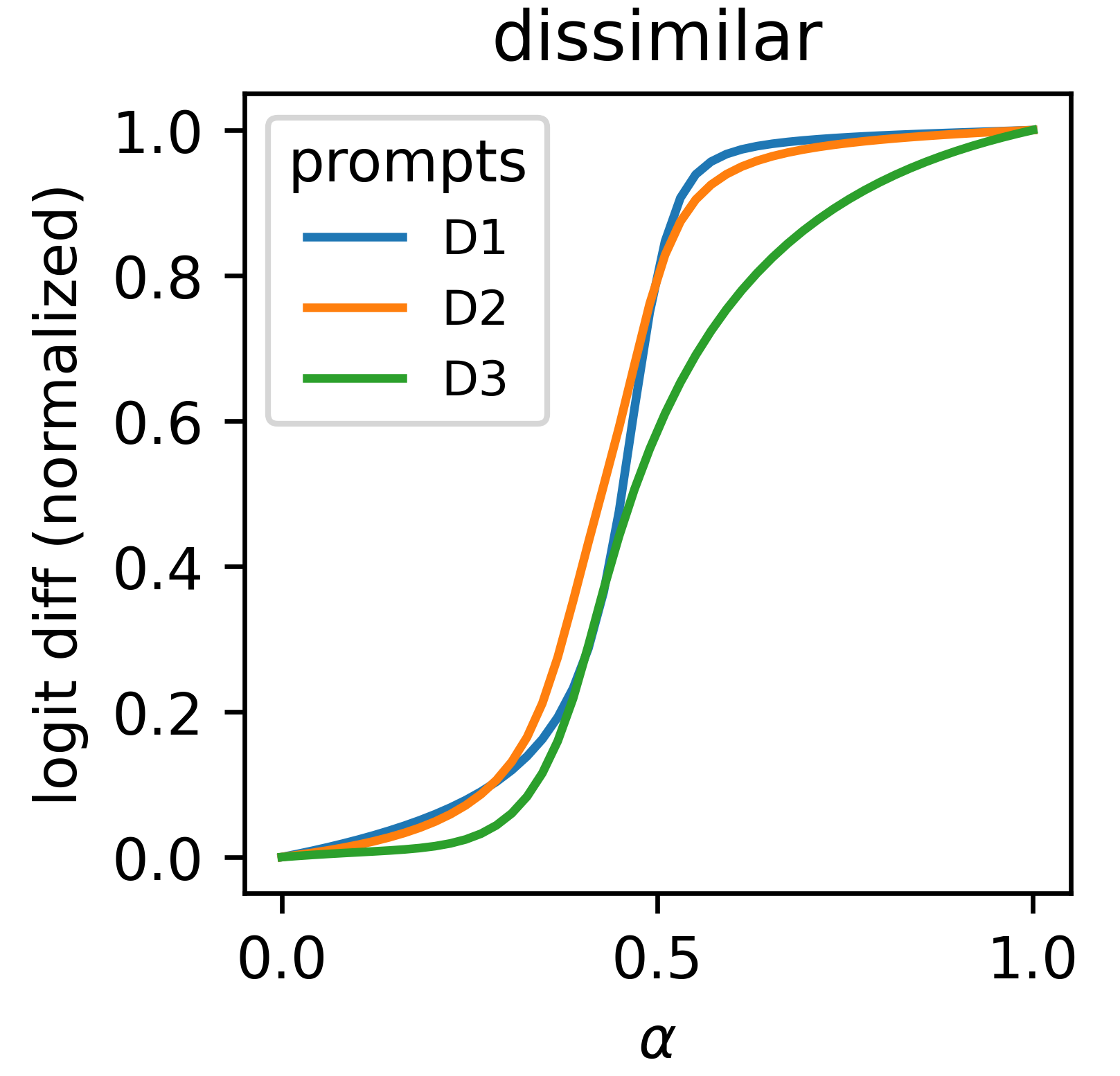}
    \caption{}\label{fig:onec}
    \end{subfigure}
    \caption{(a,b) Relative output distance as a function of $\alpha$ for (a) similar and (b) dissimilar pairs of prompts in $\texttt{Qwen2-0.5B}$. (c) Normalized logit difference between top prediction for $p_B$ and $p_A$.}%
    \label{fig:one}%
\end{figure}

In~\Cref{fig:one} we plot the relative distance $\frac{d(\alpha)}{d(1)}$ in model outputs as we interpolate between two prompts in $\texttt{Qwen2-0.5B}$\footnote{For bigger models from the Qwen2 family, we observe smaller difference between semantically similar and different prompts. We hypothesize that more capable models are able to make more precise predictions, even for very similar prompts.}. In~\Cref{fig:onea}, we show the results for interpolation between similar pairs S1-S3, and in~\Cref{fig:oneb} for dissimilar pairs D1-D3. We observe that the shapes for similar prompts S1-S3 are close to linear, or equivalently that the rate of change is similar throughout the interpolation. This is in contrast to results for dissimilar pairs D1-D3, where we observe flat regions at the beginning and at the end of the interpolations, separated by a sharp jump, which is in line with what~\citet{heimersheim_activation_2024} observed in $\texttt{GPT-2}$. This relationship between semantic similarity and stable regions is further supported by a large-scale analysis of prompt pairs filtered by their maximum sensitivity (\Cref{appendixG}). Our interpretation of these results is:
\begin{enumerate}
    \item  Activations for dissimilar prompts appear to occupy distinct stable regions. This is evidenced by the non-linear shape of $d(\alpha)$, which shows flat areas near the model-generated activations and a sharp jump between them.
    \item Activations for similar prompts likely belong to the same stable region, as indicated by the linear change in $d(\alpha)$..
    \item Stable regions are large compared to the polytopes described by \citet{black_interpreting_2022}. When interpolating between semantically different prompts, we appear to cross only a single stable region boundary, while hundreds or thousands of gates change their activation sign during the same interpolation (\Cref{appendixF}).

\end{enumerate}

Additionally, in~\Cref{fig:onec}, we plot the logit difference, normalized to $0$-$1$ range, between top predictions for both prompts in D1-D3 pairs. The change in logit difference follows a similar dynamic to $d(\alpha)$, suggesting that change in $d(\alpha)$ reflects an interpretable change in model predictions.

\paragraph{Impact of model size}

We want to test if the results we report in the previous section generalize to other prompts and models, and whether the model size plays a role. We sample 1,000 pairs of 10-token-long, unrelated prompts ($p_A, p_B$) from the $\texttt{sedthh/gutenberg\_english}$~\citep{nagyfi_sedthhgutenberg_english_2024} dataset\footnote{We chose this popular dataset of English books, as we expect it was used in training of most Transformer LMs. Additionally, we wanted to have a dataset of a limited diversity, and easily understandable next token predictions, but this will only become relevant in future work.}, and we use the methodology described for the previous experiment.

\begin{figure}%
    \centering
    \begin{subfigure}{0.3361\textwidth}
    \includegraphics[width=\textwidth]{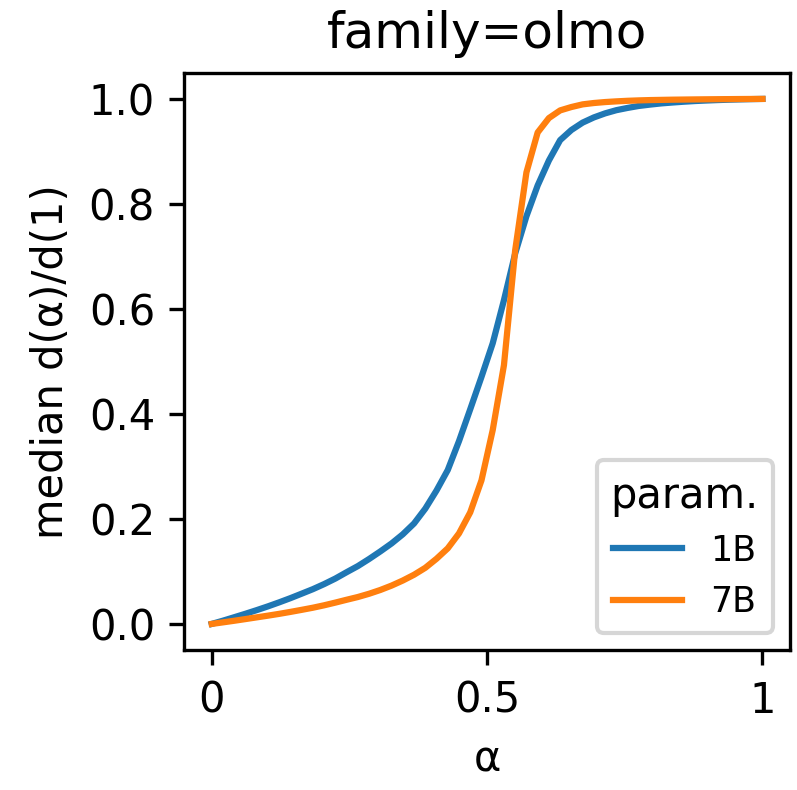}
    \caption{}\label{fig:twoa}
    \end{subfigure}
    \begin{subfigure}{0.2695\textwidth}
    \includegraphics[width=\textwidth]{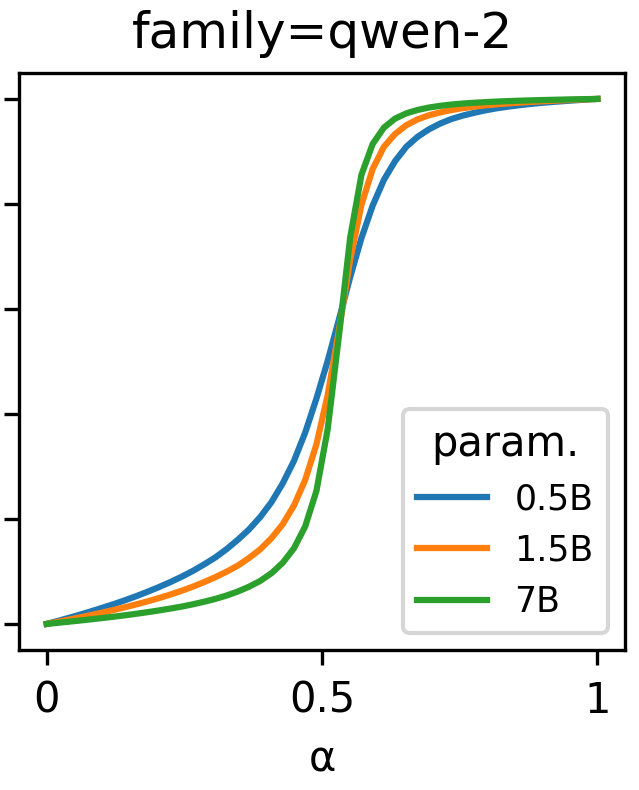}
    \caption{}\label{fig:twob}
    \end{subfigure}
    \begin{subfigure}{0.3344\textwidth}
    \includegraphics[width=\textwidth]{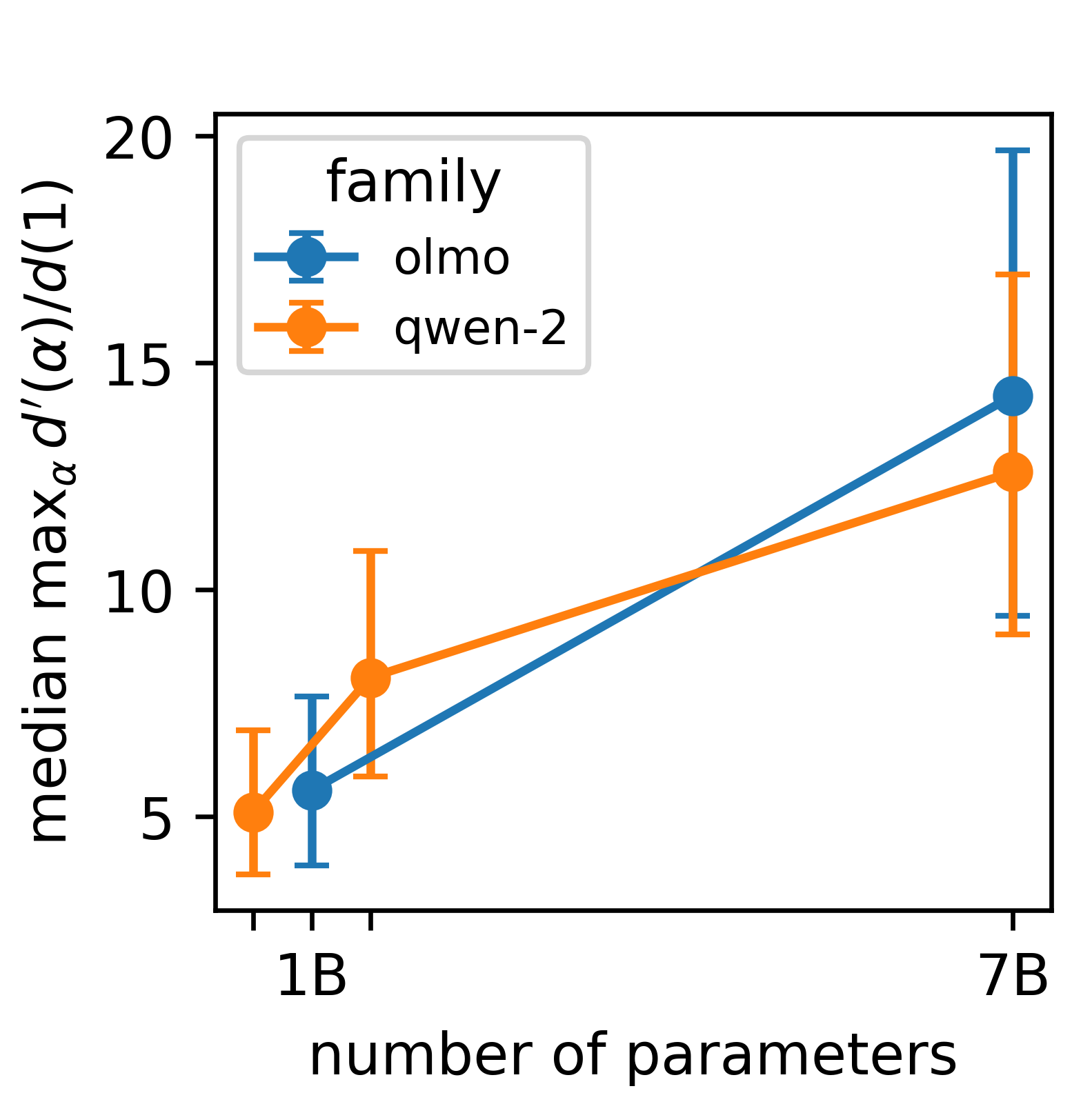}
    \caption{}\label{fig:twoc}
    \end{subfigure}
    \caption{(a,b) Median relative output distance as a function of interpolation coefficient $\alpha$ for different models from the (a) $\texttt{OLMo}$ and (b) $\texttt{Qwen2}$ families. (c) Maximum slope as a function of the number of parameters for both model families. Dots represent median, and error bars represent 25th and 75th percentiles.}%
    \label{fig:two}%
\end{figure}

In~\Cref{fig:twoa,fig:twob} we plot the median relative distance for different model sizes from $\texttt{OLMo}$ and $\texttt{Qwen2}$ families respectively. The curves appear to get sharper with increased model size. We quantify this sharpening effect in~\Cref{fig:twoc} by plotting the median of maximum slope of the relative distance, which confirms the qualitative observation.

\paragraph{Impact of training progress}

\begin{figure}%
    \centering
    \begin{subfigure}{0.3302\textwidth}
    \includegraphics[width=\textwidth]{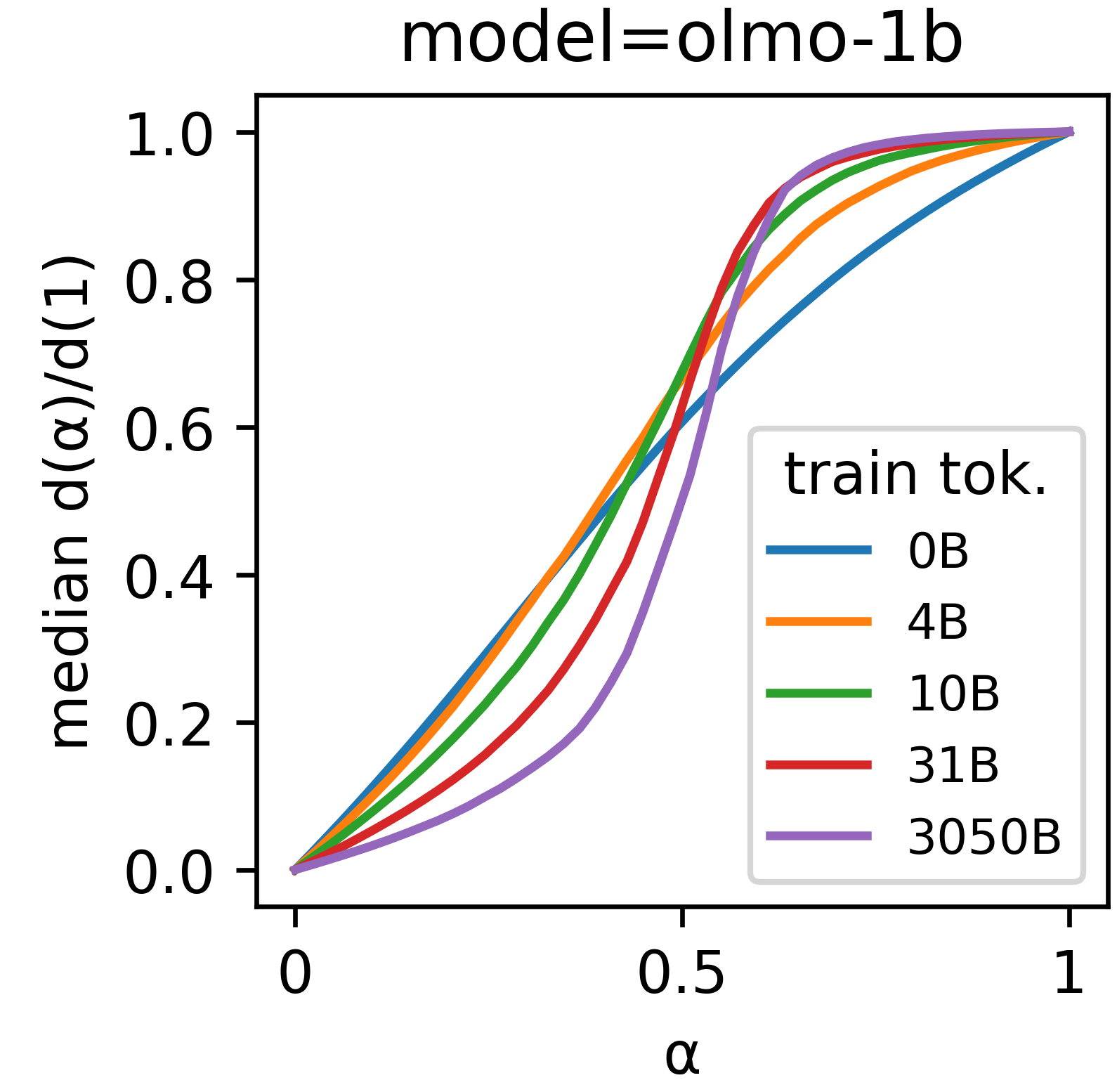}
    \caption{}\label{fig:threea}
    \end{subfigure}
    \begin{subfigure}{0.2629\textwidth}
    \includegraphics[width=\textwidth]{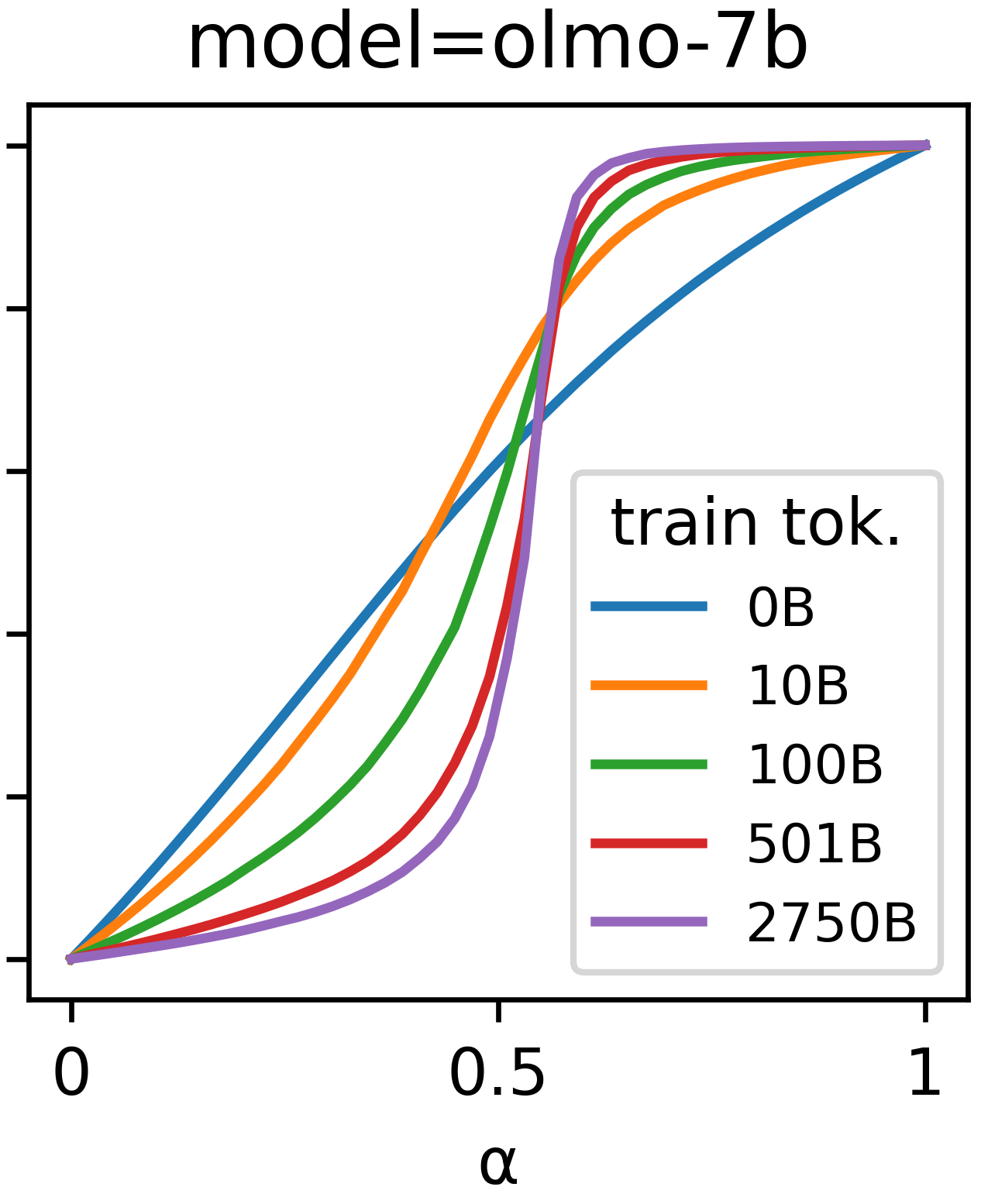}
    \caption{}\label{fig:threeb}
    \end{subfigure}
    \begin{subfigure}{0.3468\textwidth}
    \includegraphics[width=\textwidth]{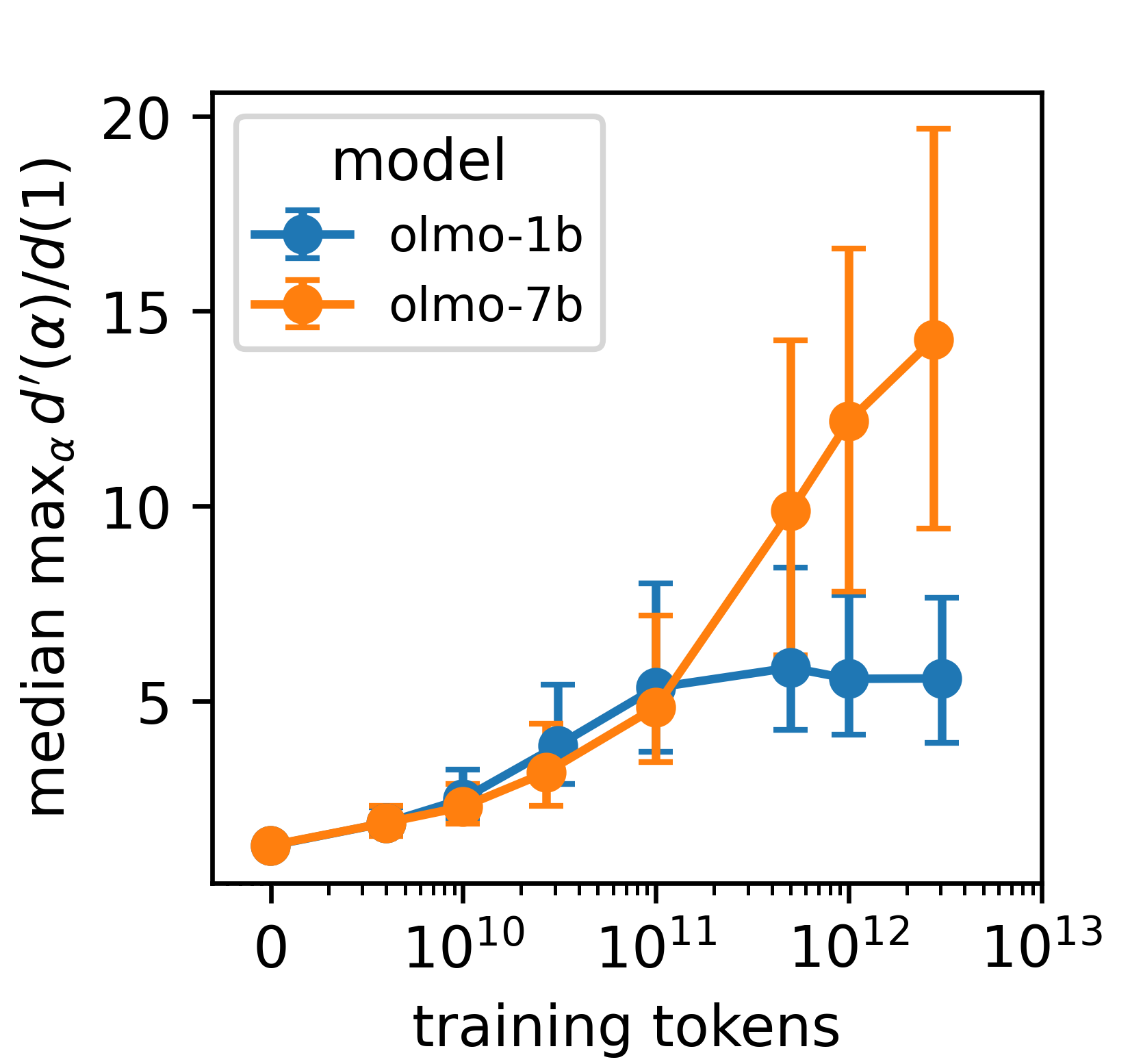}
    \caption{}\label{fig:threec}
    \end{subfigure}
    \caption{(a,b) Median relative output distance as a function of $\alpha$ for (a) $\texttt{OLMo-1B}$ and (b) $\texttt{OLMo-7B}$ models. (c) Maximum slope as a function of the number of training tokens for both models. Dots represent median, and error bars represent 25th and 75th percentiles. Note that in subfigures (a) and (b) we only show a few selected checkpoints for readability. }%
    \label{fig:three}%
\end{figure}

We repeat the experiments from the previous section, this time varying the number of training tokens instead of model size. Results presented in~\Cref{fig:three} reveal several insights. First, curves for randomly initialized models appear close to linear. However, as training progresses, we observe an increasing sharpness in the curves, similar to the effect seen with larger model sizes. This suggests the emergence and refinement of stable regions throughout the training process. Interestingly, the rate of this sharpening effect appears to plateau earlier for smaller \texttt{OLMo-1B} model compared to the larger \texttt{OLMo-7B}. 

At least two different factors may contribute to the sharpening: (1) models develop additional stable regions; (2) the boundaries between existing regions become more defined.\footnote{Factor 1 would result in more individual curves being sharp rather than linear, so the aggregated result would appear sharper. Another factor could be that the location of sharp jump is initially spread for different curves, but becomes more concentrated. That would affect the qualitative plots, (a,b) but not the quantitative plot (c).} To investigate the role of these factors, we examine a 2D slice of the residual stream after the first layer in \Cref{fig:newone}. This slice is spanned by three model-generated activations $(A, B, C)$, corresponding to the red, green, and blue circles respectively. We create synthetic activations within this slice. For each synthetic activation, we compute how similar the model's output is to the outputs produced by A, B, and C. The RGB color of each point in the plot represents these similarities: red, green, and blue components correspond to similarities with A, B, and C respectively. See \Cref{appendixC} for details.

Solid colors indicate stable regions, while color transitions represent boundaries between them. Initially, the plot shows a smooth gradient of colors. As training progresses, distinct regions of solid color emerge, and the boundaries between these regions become increasingly sharp. In \Cref{appendixD}, we present more slices for different triples of model-generated activations. Some of them hint at existing stable regions splitting during training, in particular \Cref{fig:slice_split}.

\section{Discussion}

We only look at a handful of relatively small Transformer language models. We have seen similar results for GPT, Pythia, Phi, and Llama, with a notable exception of Gemma.

We only present indirect evidence for the relatively large size of stable regions, compared to polytopes. Estimating their size or number directly, and how it changes during training and with model size, could provide a more solid ground for this claim.

\section{Acknowledgements}
This research was conducted as part of the LASR Labs program. Jacek Karwowski was supported by a grant from Open Philanthropy. We would like to thank
Lawrence Chan,
Daniel Tan,
Charlie Griffin,
Jake Mendel,
Nix Goldowsky-Dill,
Bilal Chughtai, and
James Wilken-Smith
for feedback, and Erin Robertson for guidance.

\bibliographystyle{icml2024}
\bibliography{references}

\appendix
\section{Comparison of OLMo and Qwen2 model families}\label{appendixA}
\begin{table}[H]
\centering
\begin{tabular}{|l|c|c|c|c|c|}
\hline
\textbf{Feature} & \textbf{\texttt{OLMo-1B}} & \textbf{\texttt{OLMo-7B}} & \textbf{\texttt{Qwen2-0.5B}} & \textbf{\texttt{Qwen2-1.5B}} & \textbf{\texttt{Qwen2-7B}} \\
\hline
Hidden Size & 2048 & 4096 & 896 & 1536 & 3584 \\
\hline
\# Layers & 16 & 32 & 24 & 28 & 28 \\
\hline
\# Attention Heads & 16 & 32 & 14 & 12 & 28 \\
\hline
\# Key/Value Heads & 16 & 32 & 2 & 2 & 4 \\
\hline
Intermediate Size & 8192 & 11008 & 4864 & 8960 & 18944 \\
\hline
Normalization & \multicolumn{2}{c|}{Non-parametric layer norm} & \multicolumn{3}{c|}{RMSNorm} \\
\hline
Vocabulary Size & \multicolumn{2}{c|}{50,304} & \multicolumn{3}{c|}{151,936} \\
\hline
Weight Tying & Yes & No & Yes & Yes & No \\
\hline
\end{tabular}%
\caption{Comparison of OLMo and Qwen2 model configurations}
\label{tab:olmo-qwen2-detailed-comparison}
\end{table}

\section{Similar and dissimilar prompts}\label{appendixB}
D1:\\
$p_A$ = `` The house at the end of the street was very'', 
top prediction = `` quiet''
\\
$p_B$ = `` The house at the end of the street was in'', 
top prediction = `` a''\\
D2:\\
$p_A$ = `` He suddenly looked at his watch and realized he was'', 
top prediction = `` late''\\
$p_B$ = `` He suddenly looked at his watch and realized he had'', 
top prediction = `` been''\\
D3:\\
$p_A$ = `` And then she picked up the phone to call her'', 
top prediction = `` mom''\\
$p_B$ = `` And then she picked up the phone to call him'', 
top prediction = ``.''\\
S1:\\
$p_A$ = `` She opened the dusty book and a cloud of mist''\\
$p_B$ = `` She opened the dusty book and a cloud of dust''\\
S2:\\
$p_A$ = `` In the quiet library, students flipped through pages of''\\
$p_B$ = `` In the quiet library, students flipped through pages in''\\
S3:\\
$p_A$ = `` The hiker reached the peak and admired the breathtaking''\\
$p_B$ = `` The hiker reached the peak and admired the spectacular''\\

\section{Details of the 2D slice visualization}\label{appendixC}
We create synthetic activations X in the 2D slice using the formula:
\[
 X = A + \alpha(B - A) + \beta(C-P),
\]
where P is the orthogonal projection of C onto the line containing A and B, and $\alpha$ and $\beta$ range from -0.25 to 1.25. This allows us to examine a broad area around and between the three reference activations.

To compute the colors for each point:

For each point $X$ in the 2D slice, we compute three distances:
\begin{align*}
d_A = \|F_{p_A}(A) - F_{p_A}(X)\|_2 \\
d_B = \|F_{p_B}(B) - F_{p_B}(X)\|_2 \\
d_C = \|F_{p_C}(C) - F_{p_C}(X)\|_2 
\end{align*}

where $F_p$ is defined as in \Cref{methods}. We normalize these distances by dividing by their respective maximum values over all points in the slice:
$$
\hat{d}_A = \frac{d_A}{\max_X d_A}, \;
\hat{d}_B = \frac{d_B}{\max_X d_B}, \;
\hat{d}_C = \frac{d_C}{\max_X d_C}
$$
The RGB color components for point X are then computed as: \texttt{Red} $= 1 - \hat{d}_A$, \texttt{Green} $= 1 - \hat{d}_B$, \texttt{Blue} $= 1 - \hat{d}_C$.

\section{More 2D slice plots}\label{appendixD}
\begin{figure}[H]
    \includegraphics[width=\textwidth]{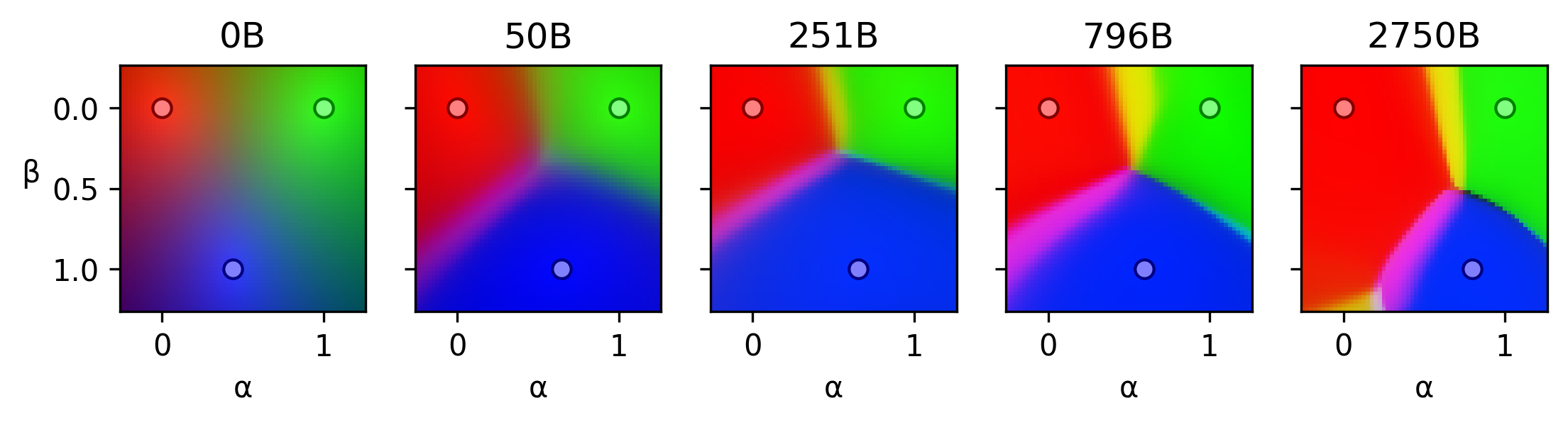}
    \caption{}
\end{figure}
\begin{figure}[H]
    \includegraphics[width=\textwidth]{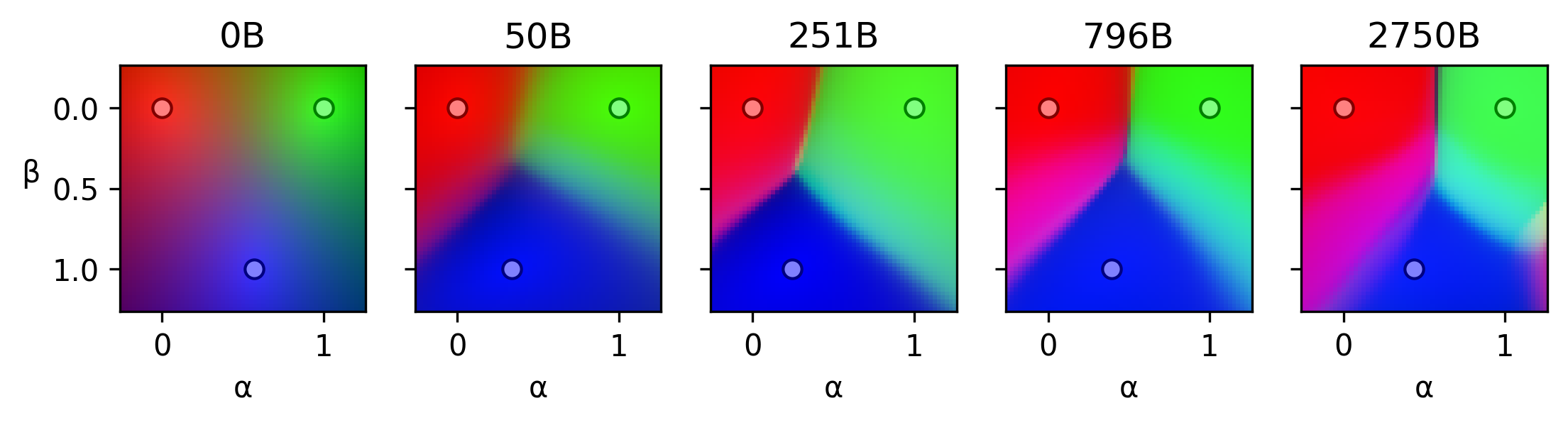}
    \caption{}
\end{figure}
\begin{figure}[H]
    \includegraphics[width=\textwidth]{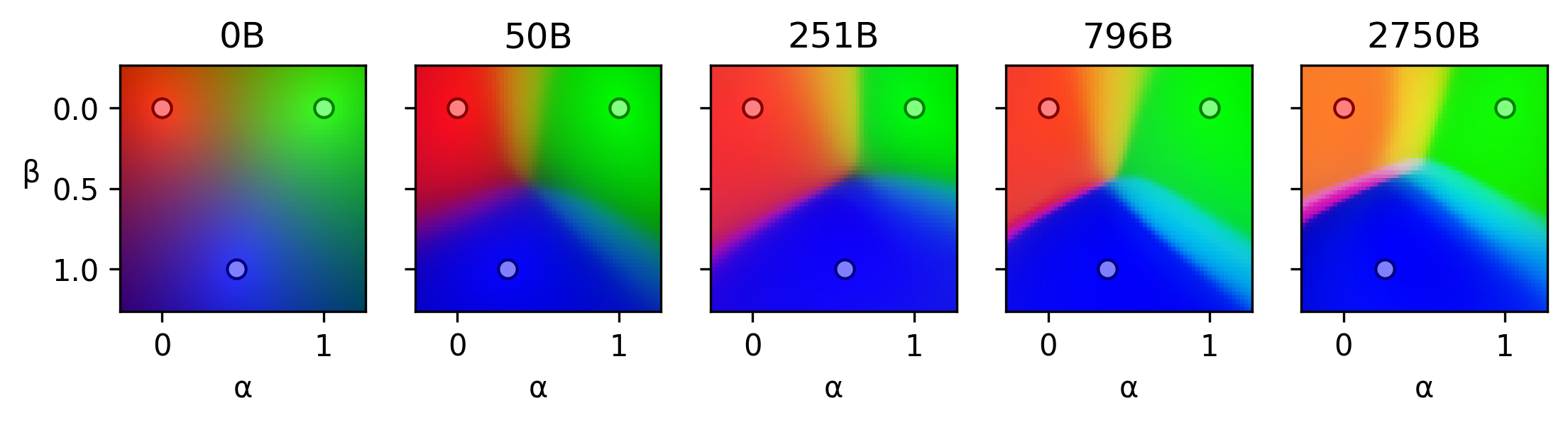}
    \caption{}
\end{figure}
\begin{figure}[H]
    \includegraphics[width=\textwidth]{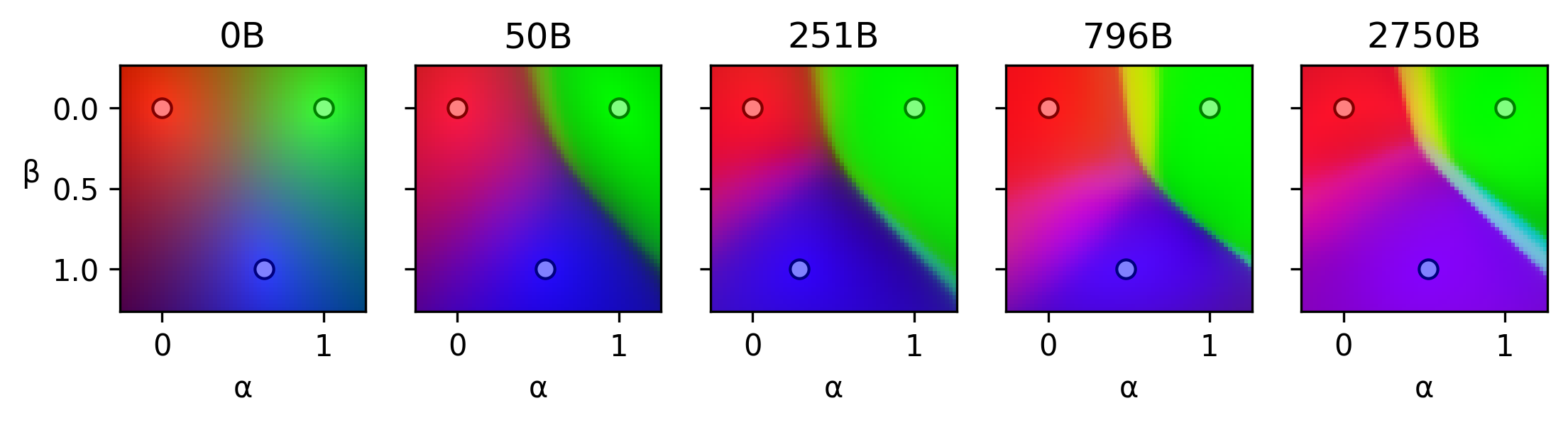}
    \caption{}
\end{figure}
\begin{figure}[H]
    \includegraphics[width=\textwidth]{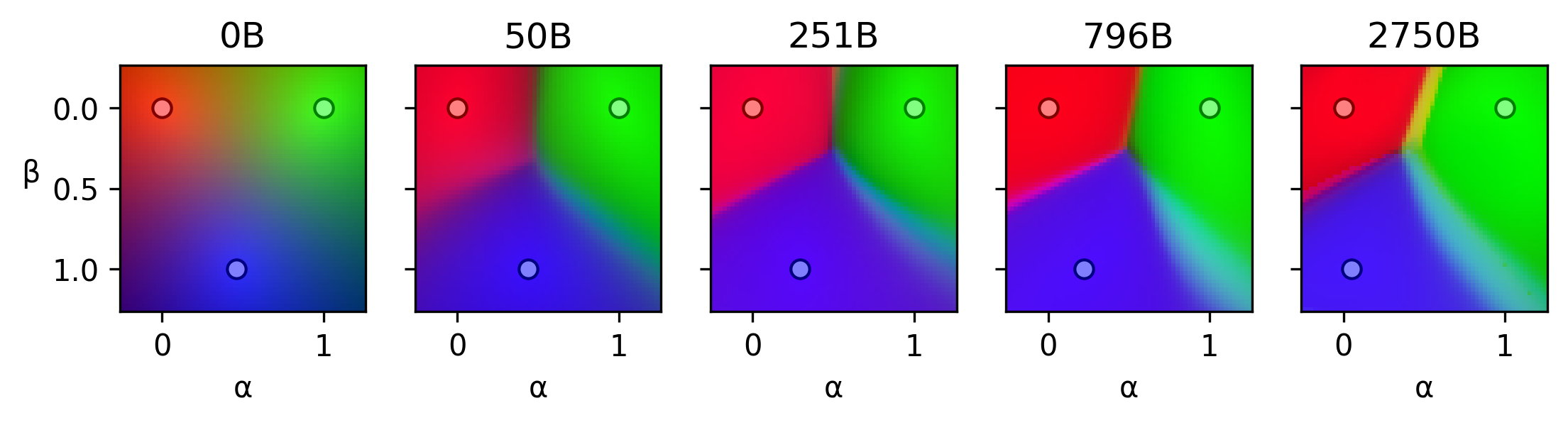}
    \caption{}
\end{figure}
\begin{figure}[H]
    \includegraphics[width=\textwidth]{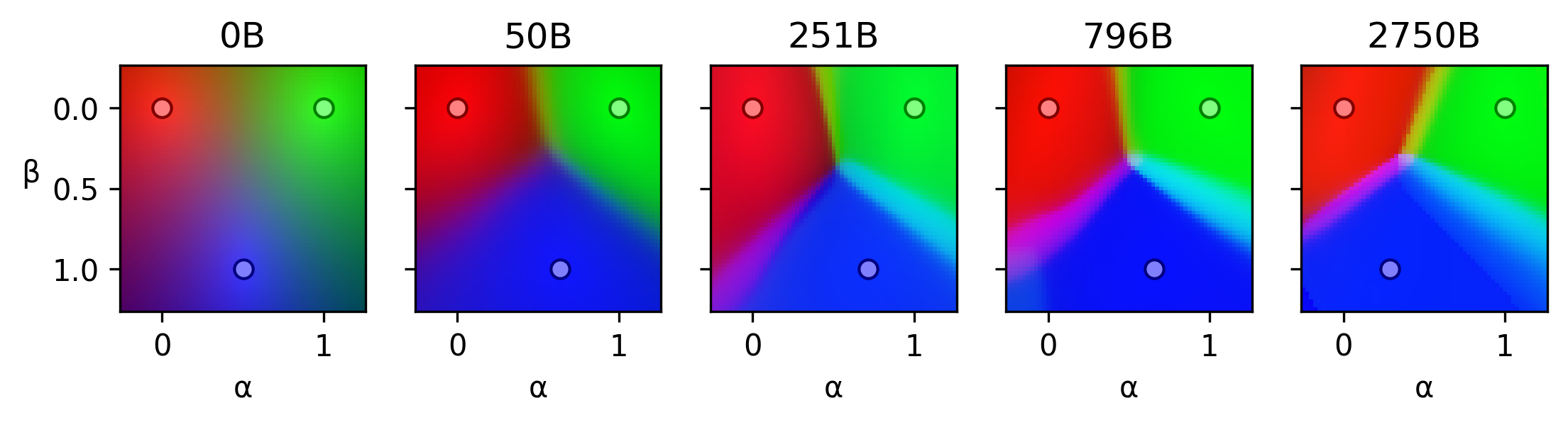}
    \caption{}
\end{figure}
\begin{figure}[H]
    \includegraphics[width=\textwidth]{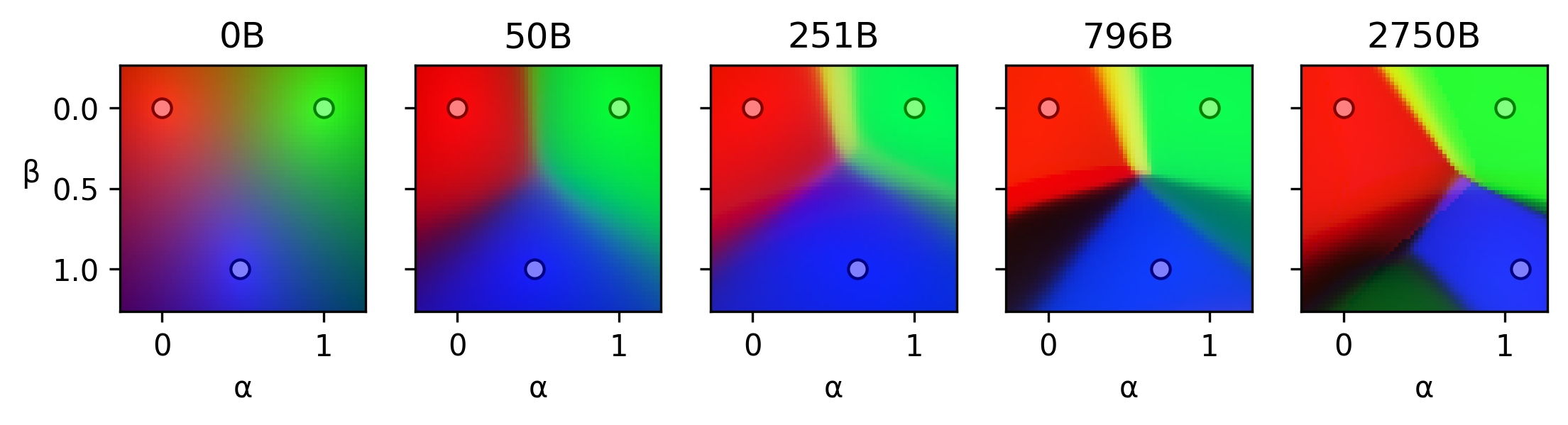}
    \caption{}
    \label{fig:slice_split}
\end{figure}
\begin{figure}[H]
    \includegraphics[width=\textwidth]{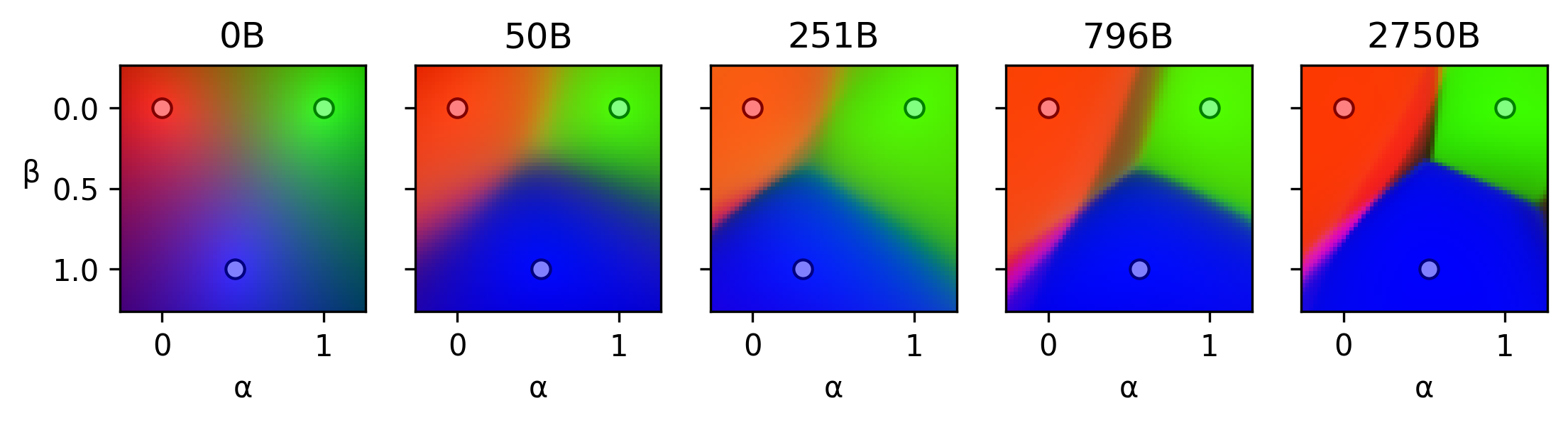}
    \caption{}
\end{figure}
\begin{figure}[H]
    \includegraphics[width=\textwidth]{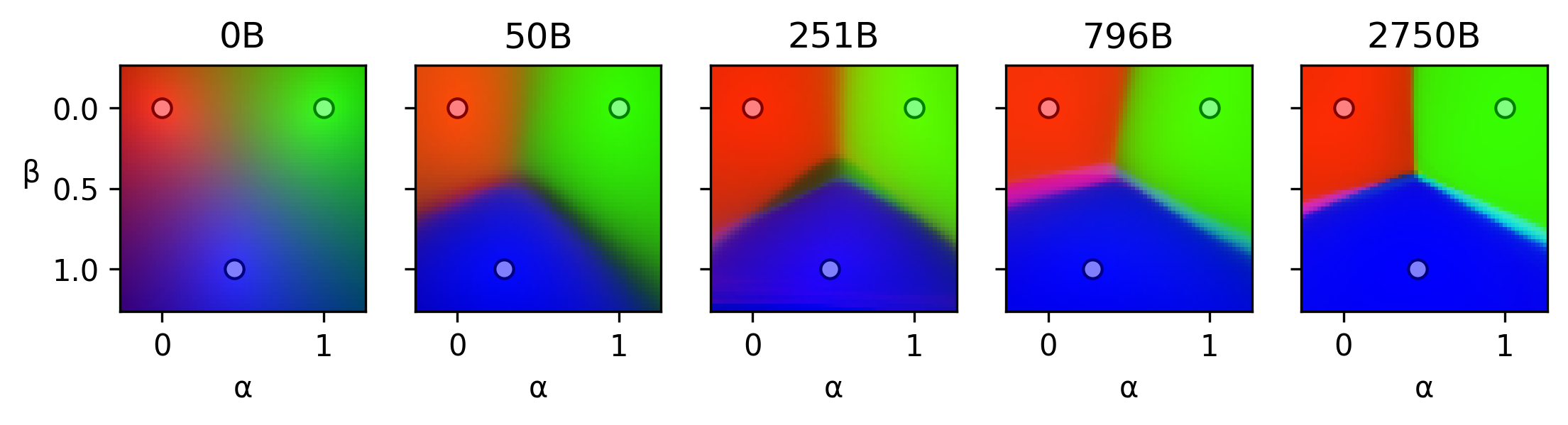}
    \caption{}
\end{figure}
\begin{figure}[H]
    \includegraphics[width=\textwidth]{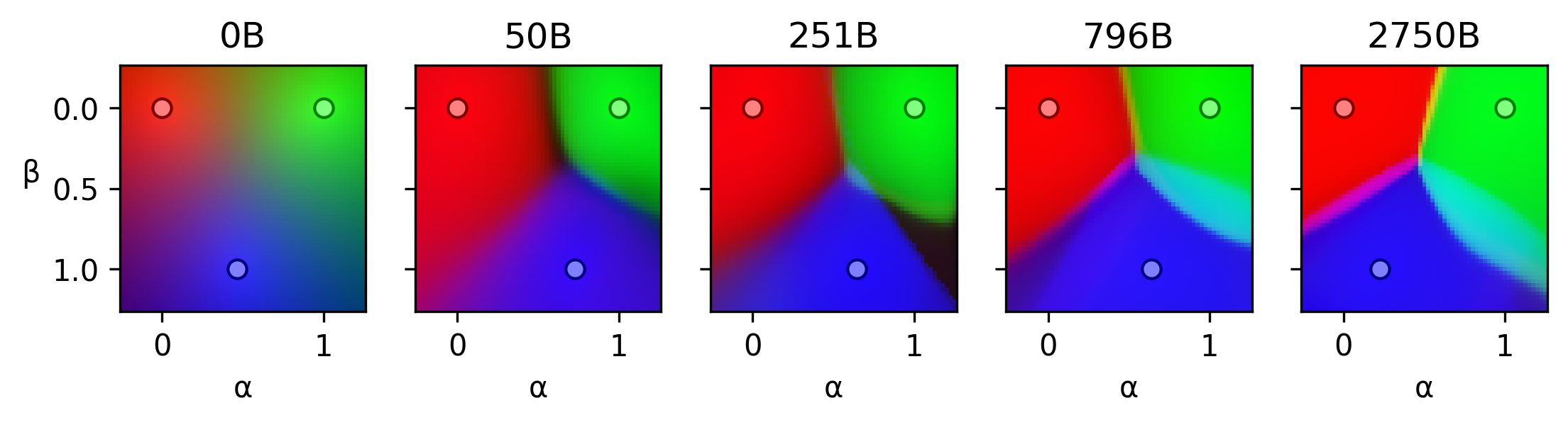}
    \caption{}
\end{figure}
\begin{figure}[H]
    \includegraphics[width=\textwidth]{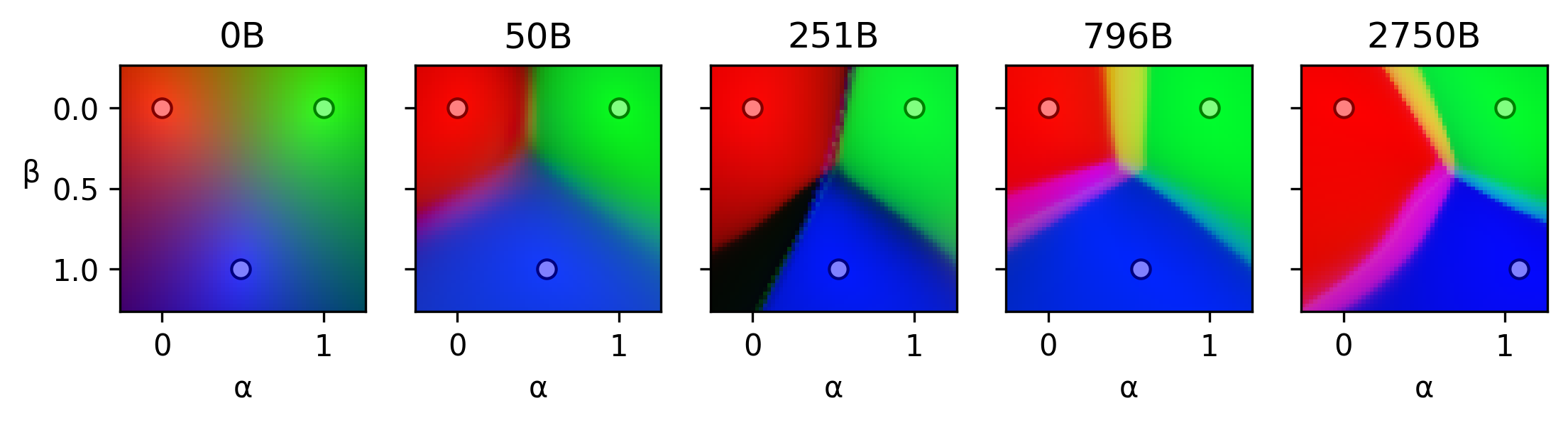}
    \caption{}
\end{figure}
\begin{figure}[H]
    \includegraphics[width=\textwidth]{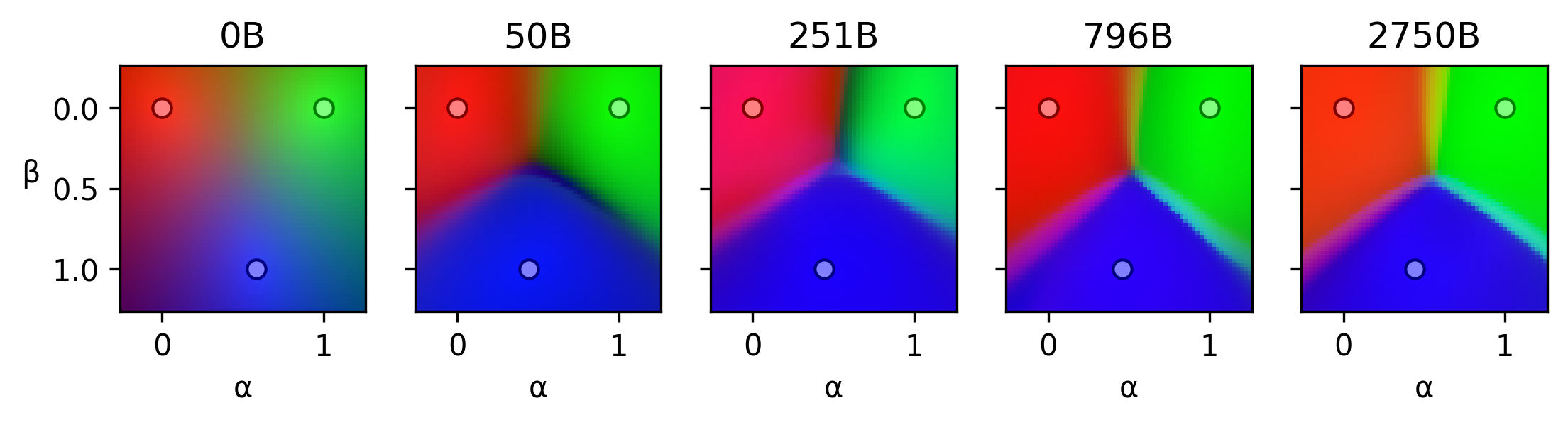}
    \caption{}
\end{figure}
\begin{figure}[H]
    \includegraphics[width=\textwidth]{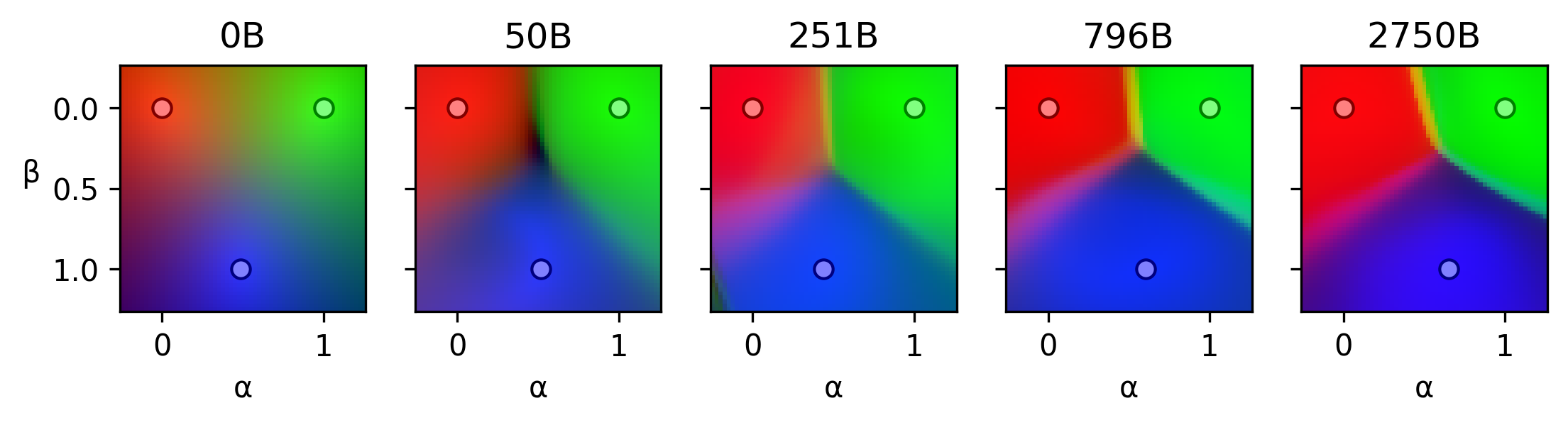}
    \caption{}
\end{figure}
\begin{figure}[H]
    \includegraphics[width=\textwidth]{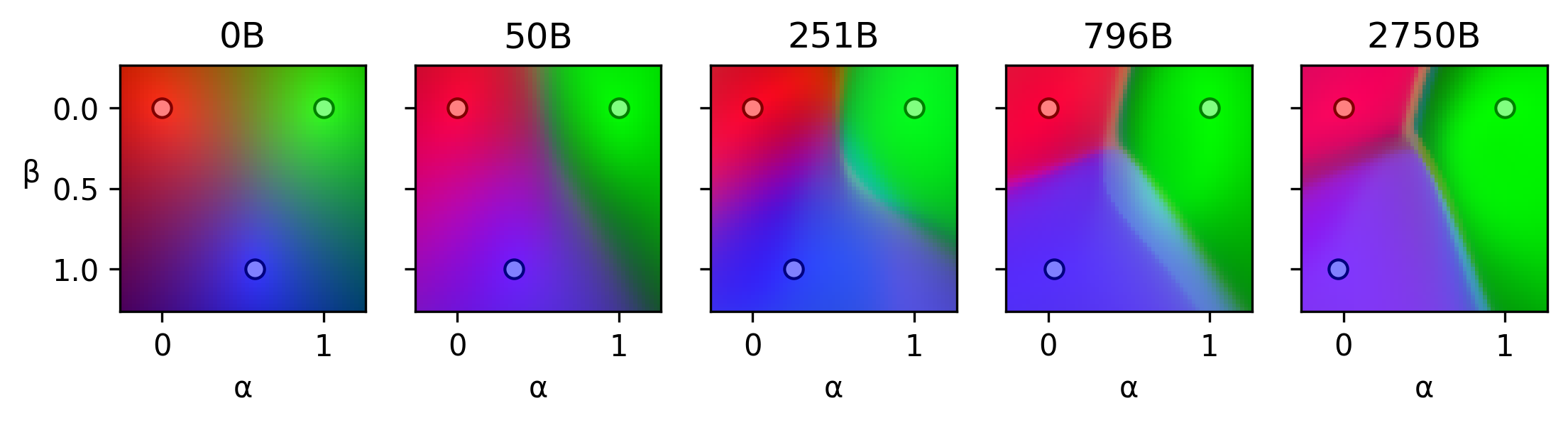}
    \caption{}
\end{figure}
\begin{figure}[H]
    \includegraphics[width=\textwidth]{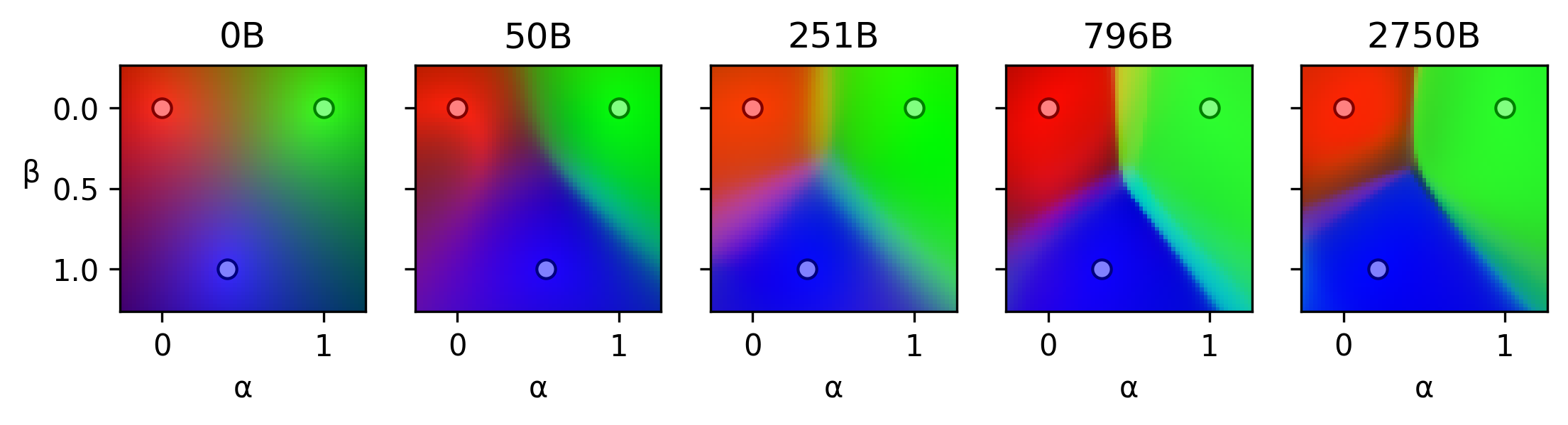}
    \caption{}
\end{figure}
\begin{figure}[H]
    \includegraphics[width=\textwidth]{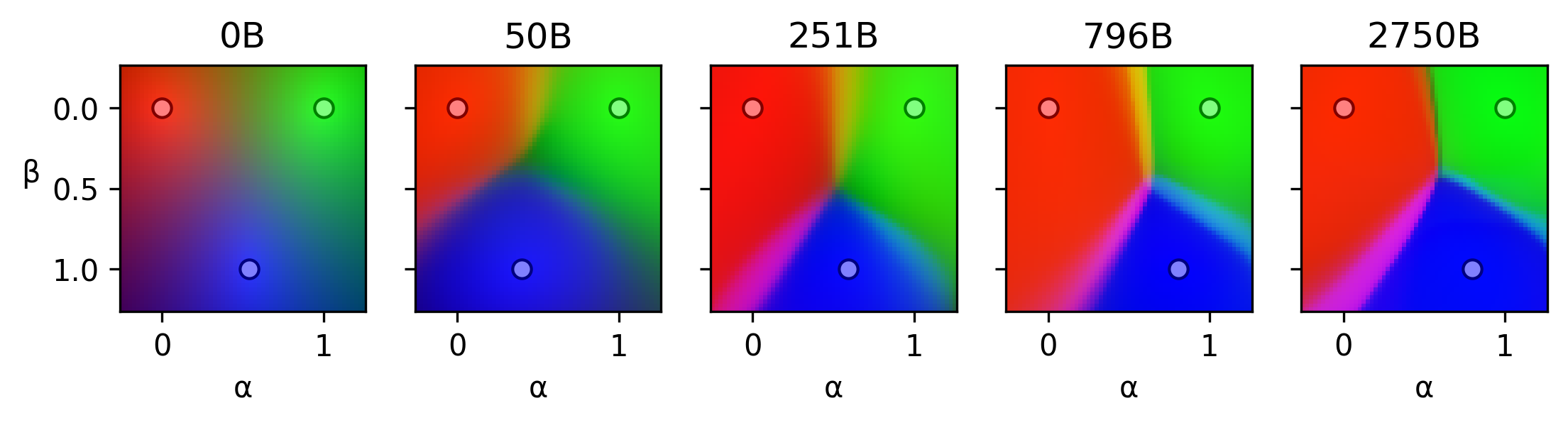}
    \caption{}
\end{figure}
\begin{figure}[H]
    \includegraphics[width=\textwidth]{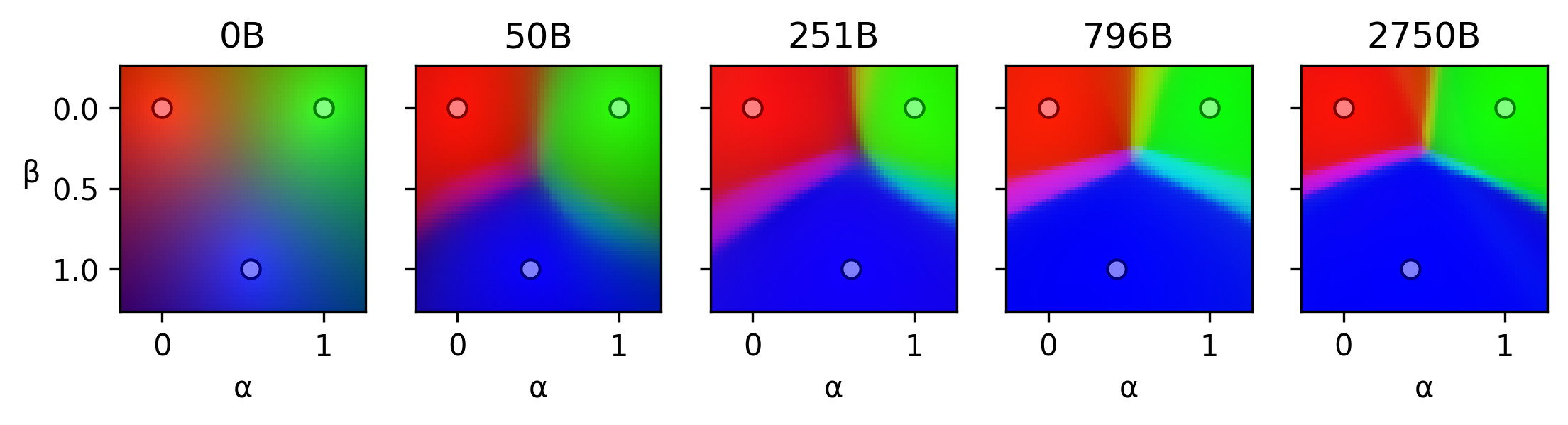}
    \caption{}
\end{figure}
\begin{figure}[H]
    \includegraphics[width=\textwidth]{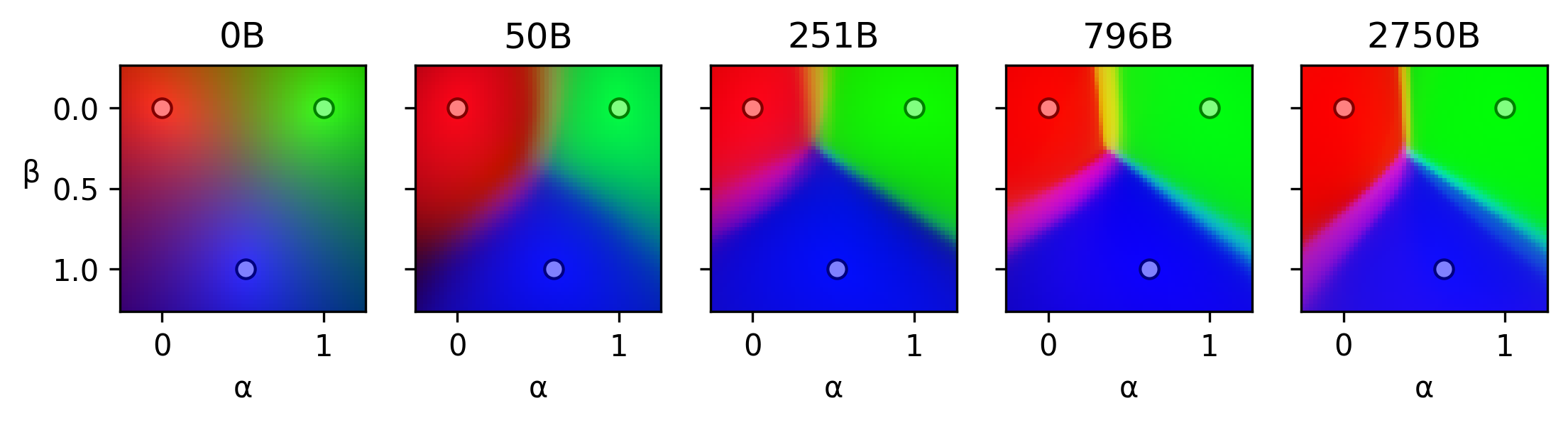}
    \caption{}
\end{figure}

\section{Shapes when patching after 1st vs after 7th layer}\label{appendixE}
\begin{figure}[H]
    \centering
    \includegraphics[width=0.6\textwidth]{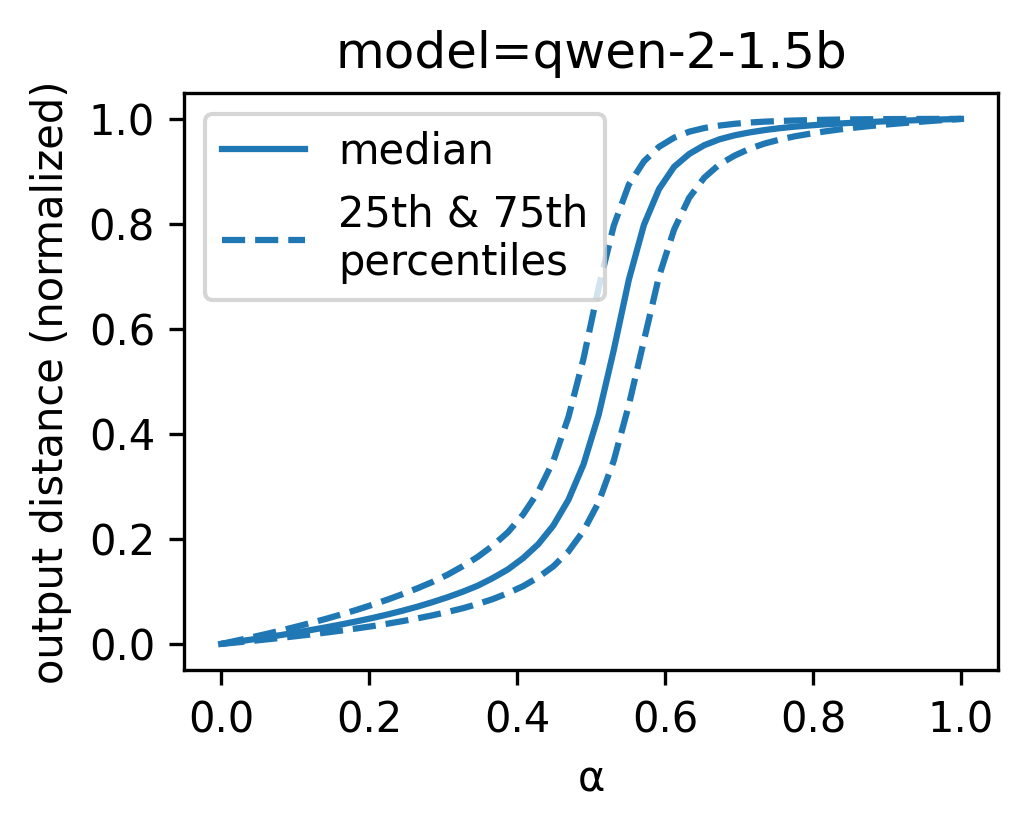}
    \caption{Interpolating and patching after 1st layer}
    \label{fig:basic_qwen_L0}
\end{figure}

\begin{figure}[H]
    \centering
    \includegraphics[width=0.6\textwidth]{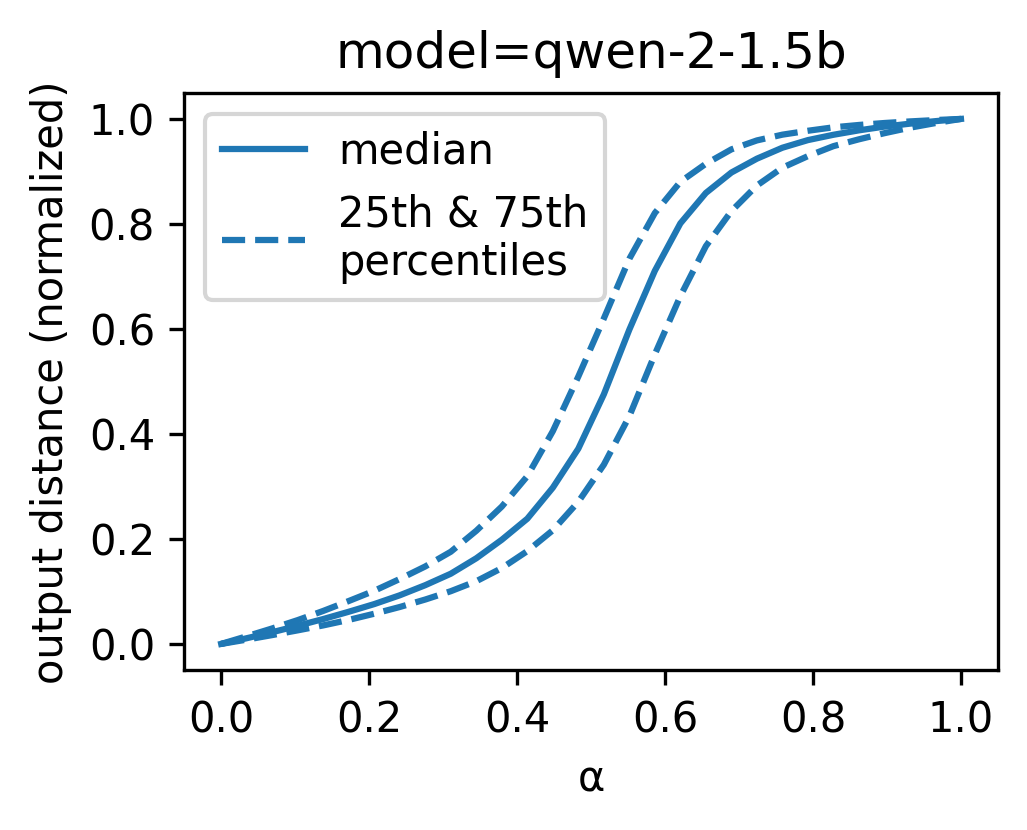}
    \caption{Interpolating and patching after 7th layer}
    \label{fig:basic_qwen_L6}
\end{figure}

We present qualitative comparison of sharpness of the shapes of $d(\alpha)$, obtained using the same methods as in \Cref{sec:experiments}, when we interpolate and patch after 1st (\Cref{fig:basic_qwen_L0}) and after 7th (\Cref{fig:basic_qwen_L6}) layer in \texttt{Qwen2-1.5B}. The y-axis is $d(\alpha)/d(1)$ in both figures. While smoother, results for layer 7 still show a significant jump around the middle of the interpolation.

\section{Direct comparison with polytopes}\label{appendixF}
To directly compare the size of stable regions with polytopes, we analyze gate activations during interpolation between prompts in \texttt{Qwen2-0.5B}. For each pair of prompts $(p_A, p_B)$, we collect gate activations at the last layer and count how many gates switch their sign (from positive to negative or vice versa) the beginning ($\alpha=0$) and the end ($\alpha=1$) of the interpolation. This provides a weak lower bound on the number of polytope boundaries crossed during interpolation.

\begin{figure}[H]
    \centering
    \includegraphics[width=0.6\textwidth]{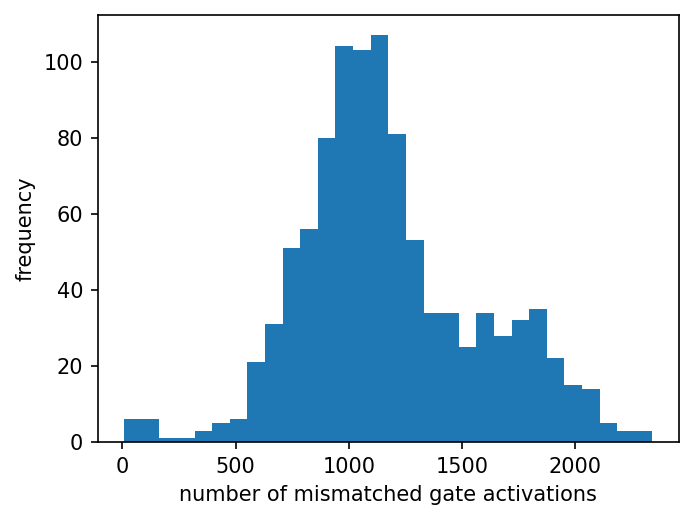}
    \caption{Histogram of the number of gate activations that switch signs during the interpolation in \texttt{Qwen2-0.5B}. Despite observing typically only one stable region boundary during interpolation, we see hundreds or thousands of gate activations changing signs, suggesting stable regions encompass many polytopes.}
    \label{fig:gates_hist}
\end{figure}

As shown in Figure~\ref{fig:gates_hist}, we typically observe hundreds or thousands of gates switching signs between two prompts, while our interpolation experiments suggest crossing only one stable region boundary. This provides direct evidence that stable regions are indeed much larger than individual polytopes.

\section{Relationship between output sensitivity and semantic similarity}\label{appendixG}

\begin{figure}[H]
    \centering
    \includegraphics[width=0.6\textwidth]{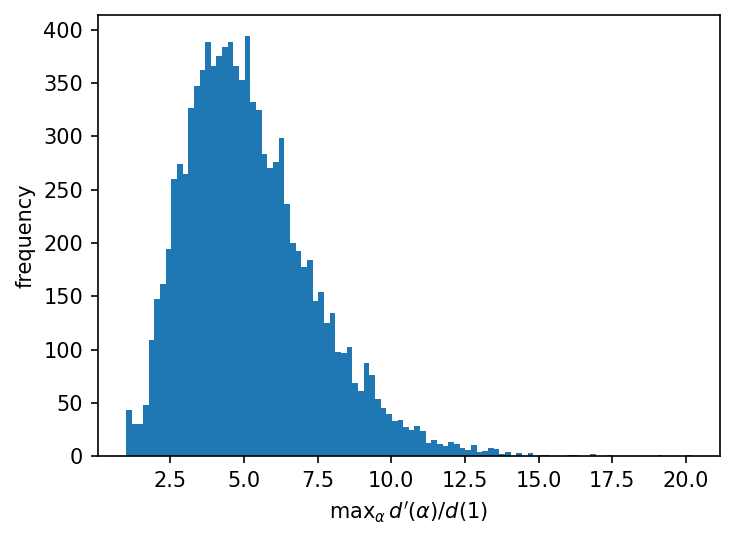}
    \caption{Distribution of maximum derivatives of the relative distance function in \texttt{Qwen2-0.5B}. The long right tail corresponds to prompt pairs that cross stable region boundaries.}
    \label{fig:max_deriv_hist}
\end{figure}

To further investigate the relationship between semantic similarity and stable regions, we analyzed prompt pairs based on the maximum derivative of the relative distance ($\max_\alpha d'(\alpha)/d(1)$) in \texttt{Qwen2-0.5B}. We found that prompt pairs with low maximum derivative ($< 1.2$) consistently share the same last token, while pairs with high maximum derivative ($> 10$) have different last tokens. This analysis provides quantitative evidence for the relationship between semantic similarity and stable regions, supporting our earlier qualitative observations with manually selected prompts.

Here are representative examples:

\subsection{Low sensitivity pairs (same last token)}

$p_A$ = ``\textbackslash{}r\textbackslash{}n\textbackslash{}r\textbackslash{}nfor the little embellishments you speak of,''\\
$p_B$ = ``ckles' arrangements with the\textbackslash{}r\textbackslash{}n\textbackslash{}r\textbackslash{}n      Angel,''\\
\\
$p_A$ = ``\textbackslash{}r\textbackslash{}n\textbackslash{}r\textbackslash{}n\textbackslash{}r\textbackslash{}nOak then struck up “Jockey to the''\\
$p_B$ = `` nothing. The sale of big furniture barely paid the''\\
\\
$p_A$ = `` being out of the question, we did the\textbackslash{}r\textbackslash{}n\textbackslash{}r\textbackslash{}n''\\
$p_B$ = ``;\textbackslash{}r\textbackslash{}n\textbackslash{}r\textbackslash{}nAnd from about him fierce effusion rolled\textbackslash{}r\textbackslash{}n\textbackslash{}r\textbackslash{}n''\\
\\
$p_A$ = ``\textbackslash{}r\textbackslash{}n\textbackslash{}r\textbackslash{}n      it had scratched the varnish from the''\\
$p_B$ = ``ically; but shall be content\textbackslash{}r\textbackslash{}n\textbackslash{}r\textbackslash{}nto produce the''\\
\\
$p_A$ = ``003:006 And when the''\\
$p_B$ = `` a year since, how much I am in the''\\
\\
$p_A$ = ``ilion\textbackslash{}r\textbackslash{}n\textbackslash{}r\textbackslash{}nand are beckoning to us.”\textbackslash{}r\textbackslash{}n\textbackslash{}r\textbackslash{}n\textbackslash{}r\textbackslash{}n''\\
$p_B$ = `` answer; but the accent was decisive enough.\textbackslash{}r\textbackslash{}n\textbackslash{}r\textbackslash{}n\textbackslash{}r\textbackslash{}n''\\
\\
$p_A$ = `` I scattered them among the heathen, and''\\
$p_B$ = `` show himself unto\textbackslash{}r\textbackslash{}n\textbackslash{}r\textbackslash{}nyou, my children, and''\\
\\
$p_A$ = ``Israel, which he commanded our fathers, that they''\\
$p_B$ = `` the LORD.\textbackslash{}r\textbackslash{}n33:31 And they''\\
\\
$p_A$ = `` quietly into the tree\textbackslash{}r\textbackslash{}n\textbackslash{}r\textbackslash{}nbeside him.\textbackslash{}r\textbackslash{}n\textbackslash{}r\textbackslash{}n\textbackslash{}r\textbackslash{}n''\\
$p_B$ = `` choir, Thea saw that she must have\textbackslash{}r\textbackslash{}n\textbackslash{}r\textbackslash{}n''\\
\\
$p_A$ = `` be Jeddak of Helium. Say\textbackslash{}r\textbackslash{}n\textbackslash{}r\textbackslash{}n''\\
$p_B$ = ``ne to hit um wid a rock? How\textbackslash{}r\textbackslash{}n\textbackslash{}r\textbackslash{}n''\\
\\
$p_A$ = `` hide the city of New York under it, and''\\
$p_B$ = ``: and their glory, and their multitude, and''\\
\\
$p_A$ = `` subterranean\textbackslash{}r\textbackslash{}n\textbackslash{}r\textbackslash{}nchambers for ages,''\\
$p_B$ = ``, to Tarzan’s\textbackslash{}r\textbackslash{}n\textbackslash{}r\textbackslash{}nconsternation,''\\
\\
$p_A$ = `` to be\textbackslash{}r\textbackslash{}n\textbackslash{}r\textbackslash{}n      entitled to his full name.''\\
$p_B$ = `` He took up his candle \textbackslash{}r\textbackslash{}n\textbackslash{}r\textbackslash{}n\textbackslash{}t\textbackslash{}t\textbackslash{}tto start.''\\
\\
$p_A$ = `` and there was the old man down the path a''\\
$p_B$ = ``as it used to be to keep from reading a''\\
\\
$p_A$ = `` more natural time, and I was once more in''\\
$p_B$ = ``.”\textbackslash{}r\textbackslash{}n\textbackslash{}r\textbackslash{}n\textbackslash{}r\textbackslash{}n“Well, there’s a good deal in''\\
\\
$p_A$ = ``, as represented by Sir Percy Blakeney,''\\
$p_B$ = `` a hundred paces,\textbackslash{}r\textbackslash{}n\textbackslash{}r\textbackslash{}nalmost without drawing breath,''\\
\\
$p_A$ = `` thee.\textbackslash{}r\textbackslash{}nPALAMON.\textbackslash{}r\textbackslash{}n\textbackslash{}r\textbackslash{}nNo more; the''\\
$p_B$ = ``      Neither boy spoke. If one moved, the''\\
\\
$p_A$ = ``GONERIL.\textbackslash{}r\textbackslash{}n\textbackslash{}r\textbackslash{}nThis admiration, sir,''\\
$p_B$ = ``: Let us go forth to meet my people,''\\
\\
$p_A$ = `` have been here a number\textbackslash{}r\textbackslash{}n\textbackslash{}r\textbackslash{}n      of times,''\\
$p_B$ = `` had laboured to do: and, behold,''\\
\\
$p_A$ = `` my things are all over Uncle Doc's house,''\\
$p_B$ = `` Paulvitch hastened back to his quarters,''\\

\subsection{High sensitivity pairs (different last tokens)}

$p_A$ = ``grandfather’s opinions for the opinions of his father''\\
$p_B$ = `` local guides:—it falls into a fatal\textbackslash{}r\textbackslash{}n\textbackslash{}r\textbackslash{}n''\\
\\
$p_A$ = `` life, but in short time\textbackslash{}r\textbackslash{}n\textbackslash{}r\textbackslash{}nAll offices of''\\
$p_B$ = `` did not know what else\textbackslash{}r\textbackslash{}n\textbackslash{}r\textbackslash{}nto do.\textbackslash{}r\textbackslash{}n\textbackslash{}r\textbackslash{}n\textbackslash{}r\textbackslash{}n''\\
\\
$p_A$ = `` conformity; as a matter of course, of course''\\
$p_B$ = `` this the emigrant Evrémonde?”\textbackslash{}r\textbackslash{}n\textbackslash{}r\textbackslash{}n''\\
\\
$p_A$ = `` kingdom\textbackslash{}r\textbackslash{}n\textbackslash{}r\textbackslash{}nalso; for he is mine elder brother''\\
$p_B$ = `` the host. And Joab and the captains\textbackslash{}r\textbackslash{}n\textbackslash{}r\textbackslash{}n''\\
\\
$p_A$ = `` view with jealousy, view with a jealous eye.\textbackslash{}r\textbackslash{}n\textbackslash{}r\textbackslash{}n''\\
$p_B$ = ``ists of\textbackslash{}r\textbackslash{}n\textbackslash{}r\textbackslash{}n      Octavius looked upon him''\\
\\
$p_A$ = ``:\textbackslash{}r\textbackslash{}n\textbackslash{}r\textbackslash{}nWell may it sort that this portentous''\\
$p_B$ = `` in time of the cholera, some people\textbackslash{}r\textbackslash{}n\textbackslash{}r\textbackslash{}n''\\
\\
$p_A$ = ``“Who are you?” asked the Scarecrow''\\
$p_B$ = `` mess of pottage, and that birthright\textbackslash{}r\textbackslash{}n\textbackslash{}r\textbackslash{}n''\\
\\
$p_A$ = ``ema. Nature, poetic, silent, balmy''\\
$p_B$ = `` way, and walk therein, and ye shall\textbackslash{}r\textbackslash{}n\textbackslash{}r\textbackslash{}n''\\
\\
$p_A$ = ``, saying, On this manner spake David.\textbackslash{}r\textbackslash{}n''\\
$p_B$ = `` medium of Clifford’s\textbackslash{}r\textbackslash{}n\textbackslash{}r\textbackslash{}nhappiness, it would''\\
\\
$p_A$ = `` before Thark, had I seen two men fight''\\
$p_B$ = ``tered. Elsewhere in the field,\textbackslash{}r\textbackslash{}n\textbackslash{}r\textbackslash{}nHere,''\\
\\
$p_A$ = `` relatively few spoken words exchanged\textbackslash{}r\textbackslash{}n\textbackslash{}r\textbackslash{}neven in long conversations''\\
$p_B$ = `` whole congregation together was forty and two thousand three\textbackslash{}r\textbackslash{}n\textbackslash{}r\textbackslash{}n''\\
\\
$p_A$ = ``ITUS LARTIUS; between them,\textbackslash{}r\textbackslash{}n\textbackslash{}r\textbackslash{}n''\\
$p_B$ = `` one of the retired\textbackslash{}r\textbackslash{}n\textbackslash{}r\textbackslash{}nstreets of a not''\\
\\
$p_A$ = ``erringly, was also manifest and indisputable''\\
$p_B$ = `` revolution\textbackslash{}r\textbackslash{}n\textbackslash{}r\textbackslash{}ncovered with its cloud this summit where r''\\
\\
$p_A$ = `` only look at her companion. Eleanor’s countenance''\\
$p_B$ = `` the tribe of Judah, the mount Zion which''\\
\\
$p_A$ = ``to go to Camden Place herself, she should not''\\
$p_B$ = `` the girl well knew, since he had been\textbackslash{}r\textbackslash{}n\textbackslash{}r\textbackslash{}n''\\
\\
$p_A$ = `` lost\textbackslash{}r\textbackslash{}n\textbackslash{}r\textbackslash{}nin thought. He retraced his steps''\\
$p_B$ = `` seemed weak and harmless.\textbackslash{}r\textbackslash{}n\textbackslash{}r\textbackslash{}n    \textbackslash{}r\textbackslash{}n\textbackslash{}r\textbackslash{}n\textbackslash{}r\textbackslash{}n      What Black''\\
\\
$p_A$ = ``    \textbackslash{}r\textbackslash{}n\textbackslash{}r\textbackslash{}n\textbackslash{}r\textbackslash{}n      “Keep it, then,” said''\\
$p_B$ = `` the champions of the Cross?”\textbackslash{}r\textbackslash{}n\textbackslash{}r\textbackslash{}n\textbackslash{}r\textbackslash{}n“To the Knights''\\
\\
$p_A$ = `` tears,\textbackslash{}r\textbackslash{}n\textbackslash{}r\textbackslash{}ntheir anguish, their terrors, their''\\
$p_B$ = `` his name be George, I’ll call him Peter''\\
\\
$p_A$ = ``\textbackslash{}xa0MERCHANT. O, 'tis a''\\
$p_B$ = `` to walk to and fro through the earth.\textbackslash{}r\textbackslash{}n\textbackslash{}r\textbackslash{}n\textbackslash{}r\textbackslash{}n''\\
\\
$p_A$ = `` made us for anything but this: to idolize''\\
$p_B$ = `` ruin.\textbackslash{}r\textbackslash{}n\textbackslash{}r\textbackslash{}n\textbackslash{}xa0\textbackslash{}xa0\textbackslash{}xa0\textbackslash{}xa0Is this your Christian counsel?''\\

\end{document}